\documentclass{amsart}

\usepackage[foot]{amsaddr}
\usepackage[T1]{fontenc}
\RequirePackage{amsthm,amsmath,amsfonts,amssymb}
\RequirePackage[authoryear]{natbib}
\RequirePackage{graphicx}
\usepackage{algorithm}
\usepackage[noend]{algpseudocode}
\usepackage{bm}
\usepackage{bbm}
\usepackage{soul}
\usepackage{float}
\usepackage{lineno}

\theoremstyle{remark}

\begin{document}

\title[Airflow recovery from THO and ABD using SST and GP]{Airflow recovery from thoracic and abdominal movements using Synchrosqueezing Transform and Locally Stationary Gaussian Process Regression}

\author{Whitney K. Huang${}^1$}
\address{${}^1$School of Mathematical and Statistical Sciences,
Clemson University}
\email{${}^1$wkhuang@clemson.edu}

\author{Yu-Min Chung${}^2$}
\address{${}^2$Department of Mathematics and Statistics,
University of North Carolina at Greensboro}
\email{${}^2$y\_chung2@uncg.edu}

\author{Yu-Bo Wang${}^3$}
\address{${}^3$School of Mathematical and Statistical Sciences,
Clemson University}
\email{${}^3$yubow@clemson.edu}

\author{Jeff E. Mandel${}^4$}
\address{${}^4$Anesthesiology \& Critical Care, Perelman School of Medicine at the University of Pennsylvania}
\email{${}^4$mandelj@uphs.upenn.edu}

\author{Hau-Tieng Wu${}^5$}
\address{${}^5$Department of Mathematics and Department of Statistical Science, Duke University, Durham, NC, USA; Mathematics Division, National Center for Theoretical Sciences, Taipei, Taiwan}
\email{${}^5$hauwu@math.duke.edu}

\maketitle

\begin{abstract}
Airflow signal encodes rich information about respiratory system. While the gold standard for measuring airflow is to use a spirometer with an occlusive seal, this is not practical for ambulatory monitoring of patients. 
Advances in sensor technology have made measurement of motion of the thorax and abdomen feasible with small inexpensive devices, but estimation of airflow from these time series is challenging. We propose to use the nonlinear-type time-frequency analysis tool, synchrosqueezing transform, to properly represent the thoracic and abdominal movement signals as the features, which are used to recover the airflow by the locally stationary Gaussian process.
We show that, using a dataset that contains respiratory signals under normal sleep conditions, an accurate prediction can be achieved by fitting the proposed model in the feature space both in the intra- and inter-subject setups. We also apply our method to a more challenging case, where subjects under general anesthesia underwent transitions from pressure support to unassisted ventilation
to further demonstrate the utility of the proposed method.    

{Keyword}:
{high-frequency physiological data};
{Gaussian process regression};
{time-frequency analysis};
{synchrosqueezing transform}.
{telemedicine}

\end{abstract}

\section{Introduction}

Breathing is an integrated physical activity involving different anatomical structures that mechanically transfer gases between alveoli and the environment. The recorded breathing activity is called the {\em respiratory (or breathing) signal}, which provide clinicians information for decision making. Numerous sensors have been described, including spirometer, piezo-sensor, electrocardiogram (ECG), photoplethysmogram (PPG), infrared video, to name but a few \citep{liu2019recent,charlton2017extraction}, to directly or indirectly acquire respiratory signals from different physiological aspects, ranging from airflow, thoracic movement (THO) and abdominal movement (ABD), impedance, physiological variability, to light spectrum. 
Sensing modalities vary in their intrusiveness, cost, and the richness of the physiological data they present.
 For example, limited respiratory data can be extracted from the ECG or PPG, including respiratory rate \citep{charlton2017extraction} and tidal volume \citep{sayadi2014optimized}, but not the expiration information. However, the expiration information can be obtained from the spirometer.
On the other hand, 
different sensors might be applicable to different monitoring scenarios.
While the gold standard for measuring airflow is to use a spirometer with an occlusive seal, using this for continuous ambulatory monitoring is not practical \citep{folke2002comparative}. Usually, it can only be carried out without the resources of a hospital. 
While recording the chest/abdominal girth for monitoring of respiratory efforts may require periodic recalibration to provide accuracy, it can be performed continuously. A system that permits accurate tracking of minute ventilation from inexpensive sensors mounted on the chest and abdomen might be useful if it provides a ``pulse oximeter for the lungs'', the airflow signal.

Motivated by the dramatically increased demands for telemedicine, for example, detection of onset of COVID-19,
opioid overdose, or asthma attacks due to worsening air pollution, it is useful to be able to extract airflow signal from 
easy-to-install sensors that are suitable for homecare. 
The signals from different sensors are in general correlated, as they are monitoring a single integrated system. 
To this end, we examine the ability to
predict airflow using ABO and THO  from the physiological perspective. Under normal physiology, the cross sectional area and longitudinal length change of the rib cage are directly related to the lung volume.
Since the THO sensor estimates the cross sectional area of the thoracic cage, and the ABD sensor estimates the increased abdominal girth due to displacement by the lung and
the diaphragm movement, 
ABD and THO together provide information about the airflow signal. If lung volume is adequately represented by ABD and THO, a model relating these should accurately predict airflow. 
However, it is  difficult to model airflow from THO and ABD due to the time-varying nonlinear nature of the physiology.

To our knowledge, there are limited studies in this direction, except \citep{mandel2016}. 
To handle this challenging problem, the authors in \cite{mandel2016} decomposed the ABD and THO signals by the Empirical Mode Decomposition (EMD) \citep{huang1998}, and demonstrated that a regression model built on the decomposed empirical modes can yield higher accuracy on predicting spirometer flow than that of the regression model with time-domain features. While the results are encouraging and exciting, however, there are several shortcomings. First, a non-trivial trial and error is needed to make EMD work, and only the in-sample predictions were carried out.
Second, how EMD maps ABD and THO to any underlying physical property of the system is unclear. Indeed, so far a theoretical support of EMD is still missing, and hence we have limited understanding of the physical meaning of the decomposed empirical modes. This fact limits a further development based on EMD from the scientific research perspective.
Due to the above limitations and its clinical importance, it is desirable if the study in \cite{mandel2016} can be extended and we could develop a theoretically solid airflow prediction model to predict the airflow signal of a subject with only ABD and THO.

To this end, we shall examine the decomposition idea considered in \cite{mandel2016}. Note that physiological signals, such as airflow, ABD and THO, often exhibit complicated oscillatory patterns, which contain crucial information about a person's health.
It is a natural idea to decompose a complicated oscillatory signal into simple ingredients as a new representation, and analyze the signal by analyzing each ingredient in this new representation. EMD is an approach considered in \cite{mandel2016} to achieve this goal.
While there are various ways to represent an oscillatory signal, some representations are more suitable than others. 
To appreciate this fact, recall that most vibrating objects have more than one {\em resonant} oscillations. Usually, a harmonic is defined as a resonant oscillation whose frequency is an integer multiple of the fundamental frequency. Physically, different harmonics play different roles and integrate to describe the characteristic vibrational mode or the {\em standing wave} pattern of the signal. In other words, harmonics encode intrinsic mechanism of vibration. Mathematically, consider a toy oscillatory signal $s(t)$, where $s$ is a $1$-periodic function (i.e., $s(t+1) = s(t),\, \forall t$). Clearly, at time $t$, the signal can be represented as $s(t)$, or equivalently as a vector $[\alpha_1\cos(2\pi t+\beta_1),\ldots,\alpha_l\cos(2\pi lt+\beta_l),\ldots,\alpha_L\cos(2\pi Lt+\beta_L)]^\top\in \mathbb{R}^L$, where $s(t)$ satisfies the Fourier series expansion $\sum_{l=1}^L \alpha_l\cos(2\pi lt+\beta_l)$. We call this vector-valued representation the {\em harmonic representation} of the oscillatory signal. The harmonic representation captures the detailed oscillatory behavior down to the harmonic level, and hence provides more information about the mechanism of vibration.
Moreover, this harmonic representation holds for signals as complicated and nonstationary as the respiratory signals (see Fig.~\ref{fig:fig1} for an example). 
We mention that the empirical modes of respiratory signals decomposed by EMD in \cite{mandel2016} are in general not harmonic representation.

Motivated by the relationship between the harmonic representation and the underlying mechanism, in this study, we propose to use Gaussian process (GP) regression with the harmonic representations as covariates to predict the flow signal. We propose to determine the harmonic representations of ABD and THO using the nonlinear-type time-frequency (TF) analysis tool, synchrosqueezing transform (SST).
We hypothesize that this approach works well during slowly evolving changes in ventilation. However, its performance of tracking ``catastrophic events'' might not be as good. 
Compared with the approach used in \cite{mandel2016}, the covariates determined by SST are interpretable with theoretical supports.

The class of GP models has been widely used in spatial and spatio-temporal statistics to interpolate spatial and space-time data \citep{cressie1993,stein1999,banerjee2014,cressie2015,wikle2019}; in design and analysis of computer experiments to build a statistical emulator of a computationally expensive physical-based simulator \citep{sacks1989,currin1991,kennedy2001,santner2003design,chen2016}; and in general regression context \citep{o1978,neal1996,williams2006} where a flexible model is needed to learn the regression function. 
The effectiveness of a GP regression model critically depends on the choice of covariates. 
We consider the recently developed nonlinear-type TF analysis tool, SST, to convert ABD and THO to harmonic representations before conducting GP regression.  
SST handles a complicated and nonstationary oscillatory time series by manipulating the phase of its short-time Fourier transform (STFT) \citep{Wu:2011Thesis} or continuous wavelet transform \citep{daubechies2011}. Its theoretical and statistical properties have been extensively studied \citep{Chen_Cheng_Wu:2014,sourisseau2019inference} and it has been applied to analyze several biomedical time series. For its application to respiratory signals, see \cite{Lin_Wu_Hsu_Wang_Huang_Huang_Lo:2016} for an automatic annotation system for sleep apnea events by analyzing ABD and THO, and \cite{wu2014using} for an instantaneous respiratory rate estimator from the ECG for patients with atrial fibrillation. We refer readers with interest to \cite{wu2020current} for a current review article.

The remainder of this paper is structured as follows. In Sec.~\ref{sec2}, we provide a background for SST and GP regression.
We mention that combining SST and GP regression to predict airflow from the ABD and THO, while making the computational load tractable by developing a locally stationary GP model (see the details in Sec.~\ref{LocGP} and Sec.~\ref{sec3}) are the main novelty of this study. 
The details of our model fitting procedure is described in Sec.~\ref{sec3}.
To assess the performance of the proposed model and methods, we apply our proposal to two real databases, which are described in Sec.~\ref{sec4}.
The setup for the prediction is also described in Sec.~\ref{sec4}. Our data analysis results are presented in Sec.~\ref{sec5}. 
We conclude with a discussion of the implications of these results and highlight some future extensions.   

\section{Mathematical Background and Model} \label{sec2}

We begin with some background for our proposed model. First, the adaptive non-harmonic model (ANHM) is introduced to represent oscillatory signals. We then present SST that we will apply to convert ABD and THO to our covariates (i.e., input features) for the regression task. Next, a quick overview of the GP models is given.  In the next section, SST and GP will be combined to carry out the predictions and the associated prediction uncertainties.

\subsection{The adaptive non-harmonic model} \label{sec2.1}
\begin{figure}[bht!]
    \centering
    \includegraphics[width=4.25in]{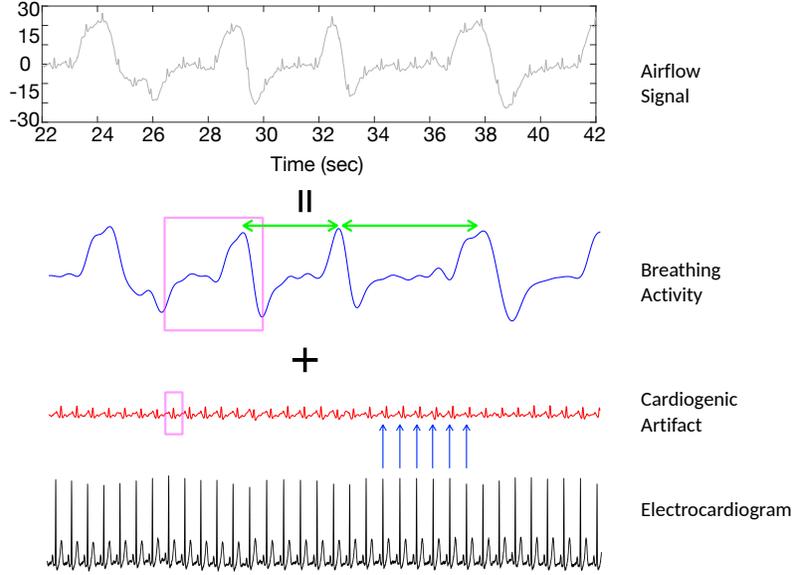}
    \caption{A typical airflow signal (gray) and its respiratory component (blue) and hemodynamic component (red). The hemodynamic component is usually called the cardiogenic artifact. The ECG signal is also shown in the bottom for a comparison (black). It is clear that some breathing cycles take a longer time and oscillate stronger than others. The two green vertical arrows indicate two cycles of different lengths. It is also clear that the oscillatory patterns of both components are not sinusoidal (magenta boxes) and vary from one to the other. Note that the cardiogenic artifact cycles coincide with the spikes shown in the ECG signal, indicated by the six blue arrows.}
    \label{fig:fig1}
\end{figure}

A primary characteristic of respiratory signals, or general biomedical signals, is oscillation. This kind of signal usually repeats a certain pattern, where the strengths (amplitude) and periods (inverse of frequency) from one cycle to the next are usually not fixed. Moreover, the oscillatory pattern is usually non-sinusoidal, and can change with time as well.
Another characteristic is that one biomedical signal might contain more than one oscillatory component. One can appreciate these features by looking at the respiratory signal of interest in this study (see Fig.~\ref{fig:fig1}). 

Due to the aforementioned nonstationarity in both amplitude and frequency, it is challenging to model respiratory signals by the commonly used time series statistical models, like the SARMA (seasonal autoregressive and moving average) model \citep{box1994,brockwe1991}, TBATS (Exponential smoothing state space model with Box-Cox transformation, ARMA errors, Trend and Seasonal components) model \citep{de2011},  seasonal-trend decomposition (STL) \citep{cleveland1990}, etc. Due to its clinical importance, the adaptive non-harmonic model (ANHM) was proposed in \cite{wu2013} to model such nonstationary oscillatory time series. Below, we review the ANHM.

Fix $\epsilon>0$. A signal $x$ satisfies the $\epsilon$-ANHM if it can be represented by
$$
x(t) = \sum_{l=1}^L A_l(t)s_l\left(\phi_l\left(t\right)\right)+\varepsilon(t)\,,
$$
where $x$ is the recorded signal, which includes $L$ oscillatory components, and the $l$-th oscillatory component, $A_l(t)s_l(\phi_l(t))$, satisfies the following conditions.
\begin{enumerate}
\item  $\phi_l$ describes the phase, which is assumed to be monotonically increasing, $C^2$, and satisfies $|\phi''_l(t)|\leq \epsilon \phi'_l(t)$ for all $t\in \mathbb{R}$. 
Since $\phi_l$ is assumed to be monotonically increasing, $\phi'_l$ is positive. It describes how fast the $l$-th component oscillates at each time.
\item $A_l(t)$ describes the amplitude of the $l$-th oscillatory component, which is assumed to be $C^1$, positive, and satisfies $|A_l'(t)|\leq \epsilon \phi'_l(t)$ for all $t\in \mathbb{R}$. 
The assumptions $|\phi''_l(t)|\leq \epsilon \phi'_l(t)$ and $|A_l'(t)|\leq \epsilon \phi'_l(t)$ for all $t\in \mathbb{R}$ indicate that the $l$-th oscillatory component $A_l(t)s_l(\phi_l(t))$ can be well approximated by $A_l(t_0)s_l[(\phi_l(t_0)-\phi'_l(t_0)t_0)+\phi'_l(t_0)t]$.
\item $s_l$ describes the ``oscillatory pattern'', which is assumed to be mean zero $1$-periodic so that its first Fourier mode is non-zero and $\|s_l\|_{L^2([0,1])}=1$.  
\item $\varepsilon$ is the noise that contaminates the signal, which is assumed to be a stationary random process with short range dependence structure. See \cite{Chen_Cheng_Wu:2014} for details. While it is possible to consider a more general noise structure, it is out of the scope of this paper, so we focus on this noise structure.
\end{enumerate}
We call $A_l(t)$ the amplitude modulation (AM), $\phi_l'(t)$ the {\em instantaneous frequency (IF)}, and  $s_l$ the {\em wave-shape function}, of the $l$-th oscillatory component. 
Take the airflow signal in Fig.~\ref{fig:fig1} as an example. One would need $L=2$ to model the THO signal. The first oscillatory component is the breathing activity, which has lower frequency and larger amplitude, and the second oscillatory component is the cardiogenic artifact, which has faster frequency and smaller amplitude.
Note that the inhalation time is shorter than the exhalation time following the normal physiology, so the oscillatory pattern $s_1$ is asymmetric, and hence it is clear that $s_1$ cannot be sinusoidal. 
For the airflow signal, the wave-shape function is usually of mean zero, since the inhaled air volume is the same as the exhaled air volume in each breathing cycle. Moreover, the oropharyngeal structure, the airway, and the environments all impact the flow pattern \citep{genta2017airflow}. So, overall the wave-shape function of the airflow signal can be viewed as an integration of the overall pulmonary system.

However, as it has been extensively discussed in \cite{lin2018}, the ANHM might not adequately describe the complicated features of physiologic signals, and we need to reconsider what the wave-shape function is. 
We start with a mathematical discussion. Take $A(t)s(\phi(t))$ as an example. By a direct Fourier series expansion, either pointwisely or in the weak sense depending on the regularity of $s$, we have
\begin{equation}\label{ANHMfirst}
A(t)s(\phi(t)) = \sum_{k=1}^{\infty}[a_{k}A(t)]\cos(2\pi k\phi(t)+\alpha_{k})\,,
\end{equation}
where $a_1>0$, $a_k\geq 0$, where $k\geq 2$, $\alpha_k\in [0,2\pi)$, $a_{1}\cos(2\pi \phi(t)+\alpha_1)$ is the {\em fundamental component}, and $a_{k}\cos(2\pi k \phi(t)+\alpha_k)$, $k\geq 1$, is the {\em $k$-th multiple} (or harmonic) of the fundamental component. In other words, a non-sinusoidal oscillation can be viewed as a summation of many sinusoidal oscillations, and the frequencies of those sinusoidal oscillations are integer multiples of the frequency of the fundamental component.
To see why the ANHM is not good enough, note that as can be visualized in Fig.~\ref{fig:fig1}, the oscillatory pattern of the airflow signal is not fixed from time to time. 
This variation could be quantified by the deviation of each multiple \citep{lin2018}. Thus, in \cite{lin2018}, it is suggested to generalize the ANHM to the following:
\begin{equation}\label{ANHMfinal}
x(t)=\sum_{l=1}^L  \sum_{k=1}^{K_l} a_{l,k}(t)\cos(2\pi \phi_{l,k}(t))+\varepsilon(t)\,,
\end{equation}
where $a_{l,k}(t)>0$ and $\phi'_{l,k}>0$ so that $|\phi'_{l,k}(t)/\phi'_{l,1}(t)-k|\leq \epsilon$. Clearly, $a_{l,k}(t)$ and $\phi_{l,k}(t)$ are generalization of $a_{k}A(t)$ and $k\phi(t)+\alpha_{k}$ in \eqref{ANHMfirst} respectively. In this work, we consider \eqref{ANHMfinal} to model different respiratory signals, and we call $\{a_{j,k}(t),\phi_{j,k}(t)\}_{k\in \mathbb{N}}$ the harmonic representation of the $j$-th oscillatory component at time $t$. Last but not the least, we emphasize that this model is not only suitable for the respiratory signal, but can also be used to model several other oscillatory time series.

\subsection{Synchrosqueezing transform (SST)}

Since harmonic representations of THO and ABD contain their key physiological features,
in the following we describe how one can use SST to extract these features. 
In short, SST is produced by applying the {\em reassignment rule} to STFT. Let us first consider STFT.
With a chosen window function $h$ (e.g., Gaussian kernel function centered at the origin), STFT is defined as
\begin{equation}
V^{(h)}_f(t, \xi) = \int f(\tau) h(\tau-t)e^{-i2\pi \xi (\tau-t)} d \tau\,,\label{eq: stft1}
\end{equation}
where $t\in\mathbb{R}$ indicates time and $\xi\in\mathbb{R}$ indicates frequency. Typically, $|V^{(h)}_f(t, \cdot)|^2$ is referred to as the {\em spectrogram} of the signal $f$ at time $t$, since it represents the power spectrum of the truncated signal $f(\cdot) h(\cdot-t)$ around $t$. 
Second, SST is defined by modifying STFT:
\begin{equation}\label{definition:SSTW}
S^{(h,\upsilon)}_{f}(t,\xi)=\int V^{(h)}_f(t,\eta) \delta_{|\xi-\Omega^{(h,\upsilon)}_f(t,\eta)|}d \eta\,,
\end{equation}
where $\xi\geq0$ and $\delta$ means the Dirac measure, and the \textit{reassignment rule} $\Omega^{(h,\upsilon)}_f$ is determined by
\begin{equation}
\Omega^{(h,\upsilon)}_f(t,\xi):=
\left\{
\begin{array}{ll}
-\Im\frac{V_f^{(\mathcal{D}h)}(t,\xi)}{2\pi V_f^{(h)}(t,\xi)}&\mbox{ when }|V_f^{(h)}(t,\xi)|> \upsilon\\
-\infty&\mbox{ when }|V_f^{(h)}(t,\xi)|\leq \upsilon.
\end{array}
\right.
\,\label{RM:omega}
\end{equation}
Here, $\mathcal{D}h(t)$ is the derivative of the chosen window function $h$, $\Im$ means the imaginary part, and $\upsilon>0$ gives a threshold so as to avoid instability in the computation when $|V^{(h)}_f(t,\xi)|$ is small.  It has been well known that $|S^{(h,\upsilon)}_{f}(t,\cdot)|^2$ is a sharpened spectrogram of the oscillatory signal at time $t$. 
Applying SST, the IF of each oscillatory component can be estimated, and each oscillatory component can be reconstructed via  \citep[Theorem 2.3.14]{Wu:2011Thesis}: 
\begin{equation}\label{eq:SST_recon_formula}
    \frac{1}{g(0)}\int_{\phi_k'(t)-\epsilon^{1/3}}^{\phi_k'(t)+\epsilon^{1/3}}S^{(h,\upsilon)}_f(t,\eta)d\eta=A(t)e^{i2\pi\phi_k(t)}+O(\epsilon).
\end{equation}
As a result, with the estimated IF, $\phi'_{1,k}$, of the respiratory signal, one is able to recover $a_{1,k}(t)\cos\left(2\pi \phi_{1,k}(t)\right)$, and hence the desired harmonic representation, $\{a_{1,k}(t),\phi_{1,k}(t)\}$. 
We refer readers to a non-mathematical tutorial \citep{wu2020new} and \citep{wu2020current} for a recent review summarizing its clinical applications.

In practice, the continuous signal $f$ is regularly sampled over a discrete set of time points with sampling interval $\Delta_t>0$. The sampling rate is hence $f_s = \Delta_t^{-1}$. Suppose the recording starts at time $t = 0$.
Write the uniformly sampled signal as a column vector $\mathbf{f}  \in \mathbb{R}^N$, where $N$ is the number of sample points and the $\ell$-th entry of $\mathbf{f}$ is $f(\ell\Delta_t)$. Take $M$ so that $2M$ is the chosen number of pixels in the frequency axis of the TF representation. 
The TF representation of $\mathbf{f}$ determined by STFT, a matrix $\mathbf{V}_\mathbf{f} \in \mathbb{C}^{N \times (M+1)}$, is then evaluated by directly discretizing the STFT formula. 
The TF representation of $\mathbf{f}$ determined by SST is also evaluated by a direct discretization of the SST formula, which is denoted as $\mathbf{S}_\mathbf{f}^\upsilon\in \mathbb{C}^{N \times (M+1)}$. The IF estimation and oscillatory component reconstruction is then carried out by
\begin{equation}\label{CurveExtractionFormula}
c^*=\max_{c\in Z_{n}^{N}}\left(\sum_{m=1}^N \log\left[\frac{|\mathbf{S}_\mathbf{f}^\upsilon(c(m),m)|}{\sum_{i=1}^n\sum_{j=1}^N|\mathbf{S}_\mathbf{f}^\upsilon(j,i)|}\right] -\lambda \sum_{m=2}^N|c(m)-c(m-1)|^2\right)\,,
\end{equation}
where $Z_n=\{1,2,\ldots,M+1\}$ and $\lambda\geq 0$. Here, $c$ indicates a curve in the TFR $|\mathbf{S}_\mathbf{f}^\upsilon|\in \mathbb{R}_+^{N\times (M+1)}$, $\sum_{m=2}^N|c(m)-c(m-1)|^2$ quantifies the regularity of the extracted curve, and $\lambda$ is the penalty term controlling the regularity of the curve $c$. Based on the robustness property of SST \citep{Chen_Cheng_Wu:2014}, the extracted curve $c^*$ is a robust estimator of the IF of the strongest IMT function.
The reconstruction formula \eqref{eq:SST_recon_formula} for the $k$-th IMT function at time $t=l\Delta_t$ is discretized into:
\begin{equation}\label{eq:SST_recon_formula2}
\frac{\Delta_\xi}{g(0)} \sum_{q\in B}S^{(h)}_f(l\Delta_t,\,q\Delta_\xi)\approx A(l\Delta_t)e^{i2\pi\phi_k(l\Delta_t)},
\end{equation}
where $\Delta_\xi>0$ is the discretization bin in the frequency axis, and 
\[
B=\{q; \phi_k'(l\Delta_t)-\epsilon^{1/3}\leq q\Delta_\xi \leq \phi_k'(l\Delta_t)+\epsilon^{1/3}\}.
\]
In practice, since $\epsilon$ is usually unknown, $B$ could be chosen to be
\[
B=\{q; \phi_k'(l\Delta_t)-b\leq q\Delta_\xi \leq \phi_k'(l\Delta_t)+b\},
\]
where $b>0$ is the frequency range chosen by the practitioner. One would take $b$ sufficiently small so that $\phi'_k(l\Delta_t)-b$ and $\phi'_{k-1}(l\Delta_t)+b$ are sufficiently separately.

\begin{figure}[H]
    \centering
    \includegraphics[width=0.9\textwidth]{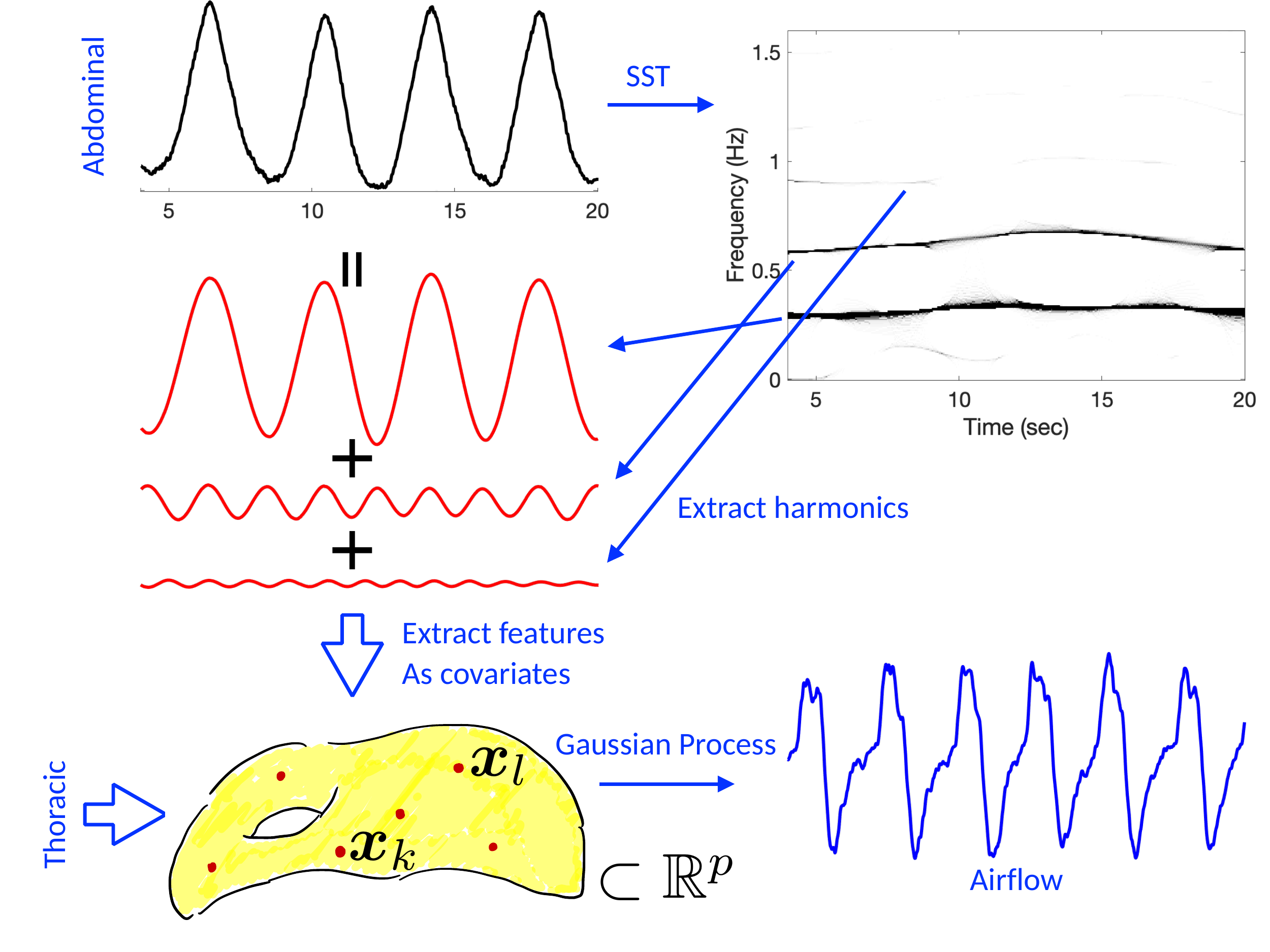}
    \caption{An illustration of the airflow recovery algorithm. The time-frequency representation (TFR) of the abdominal movement signal (black) is shown on the right-upper corner. The decomposed fundamental component and the first two harmonics are shown in red. It is clear that the amplitude of a harmonic is directly related to the intensity of its associated curve in the TFR. The extracted features, $\boldsymbol{x}_l$, from both the abdominal and thoracic movement signals are shown in the bottom, which form the input covariates of the Gaussian process to predict the airflow signal.}
    \label{fig:fig3}
\end{figure}

\subsection{Gaussian Process Regression}  \label{sec2.3}

In this work we use GP models to predict the airflow from the harmonic representations of THO and ABD (e.g., see Fig.~\ref{fig:fig3}) extracted by SST. Here a brief background on the GP model is given. First we describe the main elements of GP, namely the \textit{mean function} and \textit{covariance function}, and how these modeling elements are chosen in this work. We then discuss how we handle the computational and modeling issues when applying the GP model to long and nonstationary signals by using a {\em locally stationary} GP model. The interested reader is referred to \cite{williams2006} for a more detailed treatment on various topics of GP and its usage in machine learning and \cite{gramacy2020} for recent development in computer experiments.

\subsubsection{Nonparametric regression with Gaussian processes}

GP is a stochastic process such that every finite collection of those random variables jointly follow a multivariate normal distribution. The analytical tractability of GP makes it a powerful tool for conducting prediction and inference for a function. To fix the notation, we assume that there are $N\in \mathbb{N}$ pairs of predictor and response in the training dataset.
Let $\mathcal{Y}:=\{y_{t}\}_{t=1}^{N}$ be a set of measurements for the response of interest, which is the airflow signal in this study. The features of THO and ABD signals extracted by SST, the harmonic representations, will be treated as the covariates, and we write them as $\mathcal{X}:=\{\bm{x}_{t}\}_{t=1}^{N} \subset \mathsf{X}$, where $\mathsf{X}$ is in a finite dimensional Euclidean space. We call $\mathsf{X}$ the {\em harmonic representation space}. %
Consider the following regression model:
\begin{equation}
    y_{t} = g(\bm{x}_{t}) + \epsilon_{t}, \quad \epsilon_{t} \sim \mathrm{N}(0, \tau^2), \qquad  t = 1, 2, \cdots, N,
\end{equation}
where $g: \mathsf{X}  \mapsto \mathbb{R}$ is an unknown nonlinear function that relates the airflow to THO and ABD represented in the harmonic representation space, and $\epsilon_{t}$ is the white error term following a zero mean normal distribution with the constant variance $\tau^2$. When there is no danger of confusion, we drop the subscript $t$ in the rest of this section. GP will be used to nonparametrically model the regression function $g$ and facilitate the ultimate prediction task. From the Bayesian viewpoint, we consider a GP as a prior for the unknown regression function $g$. 

\subsubsection{Mean and covariance functions}

The specification of a GP is complete with the mean function $m(\bm{x}) = \mathbb{E}(g(\bm{x}))$, where $\bm{x} \in \mathsf{X}$, and the covariance function $\mathsf{K}(\bm{x},\bm{x}') = \mathbb{C}\text{ov}(g(\bm{x}), g(\bm{x}'))$, where $\bm{x}, \bm{x}' \in \mathsf{X}$. The common GP modeling practice is to use a relatively simple mean structure (e.g., low-order polynomial of $\bm{x}$ in spatial prediction; constant or even zero mean function in emulating computer simulations in computer experiments and non-parametric regression in machine learning) while let the covariance function absorb most local structure. 

For a given covariance function, the resulting covariance matrix for the predictors $\mathcal{X}$ is positive-definite. There exist some families of parametric covariance function that are commonly used. In general, one could also design a valid covariance function using Bochner's Theorem \cite[see][p.~208]{gikhman1976}. Additional assumptions on covariance function are typically made to facilitate the inference, specifically {\em stationarity} and {\em isotropy}. These two assumptions can be thought as the translational and rotational invariance properties so that $\mathsf{K}(\bm{x}, \bm{x}')$ only depends on the ``distance'' regardless of their ``locations'', i.e., $\mathsf{K}(\bm{x}, \bm{x}') = C(h)$, where $C$ is a valid isotropic covariance function and $h$ is an appropriate distance between $\bm{x}$ and $\bm{x}'$ in the context of interest. 

Once the mean and covariance functions have been specified, the response $\mathcal{Y}$ follows a multivariate normal distribution with the chosen mean $\bm{m} = \{m(\bm{x}_{t})\}_
{t=1}^{N}\in \mathbb{R}^{N}$ and the covariance matrix $\bm{\Sigma}_{\bm{\theta},\sigma^{2}}$, which $(i,j)$-th entry is $C_{\bm{\theta}, \sigma^2}(\|\bm{x}_{i}-\bm{x}_{j}\|)$, 
where $i,j=1,\cdots, N$, $\sigma^2$ is the marginal variance and $\bm{\theta}$ includes the parameters of the correlation function. More specifically, under the GP regression model, we have  
$$
\begin{bmatrix}
    y_{1} \\
    y_{2}  \\
    \vdots \\
    y_{N}
\end{bmatrix}
\sim \mathrm{N}\left(\begin{bmatrix}
    m(\bm{x}_{1}) \\
    m(\bm{x}_{2})  \\
    \vdots \\
    m(\bm{x}_{N})
\end{bmatrix}, 
\begin{bmatrix}
    C(\|\bm{x}_{1}- \bm{x}_{1}\|) & C(\|\bm{x}_{1}- \bm{x}_{2}\|) & \dots  & C(\|\bm{x}_{1}- \bm{x}_{N}\|) \\
    C(\|\bm{x}_{2}- \bm{x}_{1}\|) & C(\|\bm{x}_{2}- \bm{x}_{2}\|) & \dots  & C(\|\bm{x}_{2}- \bm{x}_{N}\|) \\
    \vdots & \vdots & \vdots & \vdots \\
C(\|\bm{x}_{N}- \bm{x}_{1}\|) & C(\|\bm{x}_{N}-\bm{x}_{2}\|) & \dots  & C(\|\bm{x}_{N}- \bm{x}_{N}\|)
\end{bmatrix} + \tau^2 I_{N}\right)\,. 
$$

In this study, we assume the mean function is an unknown constant in the harmonic representation space; that is, $m(\bm{x}) = \mu$ for $\bm{x} \in \mathsf{X}$.  We use the Mat\'{e}rn class covariance function \citep{stein1999}: 
$$
C(h)=\sigma^2\frac{\left(\sqrt{2\nu}h/\rho\right)^{\nu}\mathcal{K}_{\nu}\left(\sqrt{2\nu}h/\rho\right)}{\Gamma(\nu)2^{\nu-1}}, \qquad \sigma^2>0,\, \rho>0,\, \nu>0,
$$ 
where $h\geq0$, $\Gamma(\nu)$ is the gamma function, and $\mathcal K_\nu$ is the modified Bessel function of the second kind. In this work, we restrict our attention on  
a few fixed $\nu$ values ($\nu=0.5, 1.5,\infty $), which should give us some flexibility in determining the ``smoothness'' of the regression function of interest. Note that $\nu = 0.5$ and $\infty$ correspond to the exponential and squared exponential covariances, respectively.  

We use maximum likelihood method to estimate the GP parameters, $(\mu, \sigma^2, \rho, \tau^2)$. Specifically, the maximum likelihood estimates (MLE), denoted by $(\hat{\mu}, \hat{\sigma}^2, \hat{\rho}, \hat{\tau}^2)$, are obtained by finding the maximizer of the log-likelihood function:
\begin{align*}
& \ell(\mu,\rho,\sigma^2,\tau^2) = -\frac{N}{2}\log(2\pi) -\frac{1}{2}\log \left|\bm{\Sigma}+\tau^2 I\right|-\frac{1}{2}(\bm{y}-\mu\bm{1})^{\mathrm{T}}\left[\bm{\Sigma} +\tau^2 I\right]^{-1}(\bm{y}-\mu\bm{1}).
\end{align*}
For $\bm{x}_{0} \in \mathsf{X}$, a harmonic representation that may or may not be in the training set $\mathcal{X}$, $g(\bm{x}_{0})$ can be estimated using the conditional distribution formula for the multivariate normal distribution. Specifically, we use the plug-in estimator $\hat{g}(\bm{x}_{0}) = \hat{\mu} + \hat{\mathsf{k}}[\hat{\bm{\Sigma}}+\hat{\tau}^2 I]^{-1}(\bm{y} -\hat{\mu} \bm{1})$, where $\hat{\mathsf{k}} = [\hat{C}(\|\bm{x}_{0}-\bm{x}_{1}\|), \cdots, \hat{C}(\|\bm{x}_{0}-\bm{x}_{N}\|)]$. Note that this is the best linear unbiased estimator (BLUE) under the squared loss. This estimator can be expressed as a linear predictor in terms of $\bm{y}$, where the coefficients are determined by the estimated GP parameters using the same set of $\bm{y}$. Some simulation studies \citep{zimmerman1991,zimmerman1992} suggested that this ``empirical'' or ``estimated'' BLUE works well even with small $N$. Additionally, the GP model allows for quantifying the prediction uncertainty using the estimated conditional variance $\mathrm{Var}(\hat{g}(\bm{x}_{0})) = \hat{\sigma}^2 - \hat{\mathsf{k}}[\hat{\bm{\Sigma}}+\hat{\tau}^2 I]^{-1}\hat{\mathsf{k}}^{\mathrm{T}}$.

\subsubsection{Locally stationary GP} \label{LocGP}

One obstacle for the direct use of GP with the likelihood-based method is the computation of the log-likelihood function needed for parameter estimation. Specifically, the maximum likelihood estimation here requires computing the determinant and the matrix inversion repeatedly, which can be computationally prohibitive when $n$ is large due to the $\mathcal{O}(n^3)$ computation and $\mathcal{O}(n^2)$ memory storage (see \cite{heaton2019} for a recent case study for analyzing large spatial datasets and references therein or \cite{ambikasaran2015fast} for a fast method under some specific covariance structures). 
In this study, more than 200,000 data points are analyzed to predict the airflow signal. The solution we propose is fitting the GP model sequentially and ``locally'' in the harmonic representation space. Specifically, for each prediction window, we form the training data using the nearby historical observations, where these nearby observations are the nearest neighbors (NN) in terms of the Euclidean distance in the harmonic representation space. By using a small number of NN with a small prediction window in time,  the size of the training set can be controlled so that each prediction can be computed fairly quickly. 
It is worth pointing out that our proposal here is motivated by the {\em locally stationary GP} in the literature \citep{haas1990,haas1995,gramacy2015,kuusela2018}. This approach not only allows for computational tractability but also provides a means of handling possible nonstationaries.

\subsubsection{Diffusion-based GP} \label{DBGP}
Curse of dimensionality is a well-known issue in data analysis. Another commonly encountered challenge is the nonlinearity of the data structure. In our work, we encounter both challenges in the harmonic representations $\mathsf{X}$. To handle these challenges, a common approach is assuming that the covariates are supported on, or near some {\em lower-dimensional nonlinear manifold}; that is, assuming that $\mathcal{X}$ is located on a low-dimensional manifold in our work. It is thus reasonable to respect and explore such lower-dimensional nonlinear structure. 
While there have been several variations in the GP modeling context, taking the manifold structure into account is still in its infancy \citep{yang2016,calandra2016,lin2019,niu2019}. By noticing the potential of combining manifold learning techniques and the GP regression, it was recently proposed in \cite{dunson2019} to consider the symmetrized graph Laplacian (GL) as the covariance structure. Such a consideration was motivated by the fact that the GL links the random walk on the dataset to the underlying nonlinear structure via its eigenstructure.
In \cite{dunson2019}, the authors showed the spectral convergence of the GL in the $L^\infty$ sense and proposed to take the symmetrized GL to form the covariance matrix. Theoretically, it would encode the geometric and topological information in the GP regression. While this new idea hasn't been extensively examined, for a comparison purpose, we also consider this diffusion-based approach in this work.

\section{SST-Locally stationary Gaussian Process (SST-LocGP)} \label{sec3}

We now detail our model fitting procedure. There are two main steps. First, ABD and THO are represented in the harmonic representation space by SST. Then, a locally stationary GP is applied to predict the airflow signal. The proposed model is called \texttt{SST-LocGP} (\texttt{SST-LocDBGP} if diffusion-based GP is used), and it is summarized in Algorithm 1.
In the remaining of this paper, for the simplicity, we drop the prefix \texttt{SST} and refer these two models as \texttt{LocGP} and \texttt{LocDBGP}, respectively.  
 
In the first step, the THO and ABD signals are represented as \eqref{ANHMfinal}
\begin{align}
x^{(\texttt{ABD})}(t)&\,=\sum_{j=1}^2  \sum_{k=1}^{K^{(\texttt{ABD})}_j} a^{(\texttt{ABD})}_{j,k}(t)\cos\left(2\pi \phi^{(\texttt{ABD})}_{j,k}(t)\right)+\varepsilon^{(\texttt{ABD})}(t)\nonumber\\
x^{(\texttt{THO})}(t)&\,=\sum_{j=1}^2  \sum_{k=1}^{K^{(\texttt{THO})}_j} a^{(\texttt{THO})}_{j,k}(t)\cos\left(2\pi \phi^{(\texttt{THO})}_{j,k}(t)\right)+\varepsilon^{(\texttt{THO})}(t)\,,\nonumber
\end{align}
where we assume the first component, $j=1$, is the respiratory signal, and $K^{(\texttt{ABD})}_j$ and $K^{(\texttt{THO})}_j\in \mathbb{N}$ are both finite for $j=1,2$. By empirically examining the signals, it is reasonable to assume that $K^{(\texttt{ABD})}_1= 4$ and $K^{(\texttt{THO})}_1= 4$ since they give accurate approximations of the ABD and THO signals (see Fig. 1 in the Supplementary Material). 

Suppose the airflow, ABD and THO are sampled at $\xi_0$ Hz. In the preprocessing step, the local regression using weighted linear least squares and a second degree polynomial model is applied to remove the trend. 
Based on this model and the preprocessing, at each time $t_l:=l\Delta_t$, where $\Delta_t=1/\xi_0$ and $l\in \mathbb{N}$, \eqref{CurveExtractionFormula} and \eqref{eq:SST_recon_formula2} are applied to estimate $a^{(\texttt{ABD})}_{1,k}(t_l)$ and $\phi^{(\texttt{ABD})}_{1,k}(t_l), \, k=1,\ldots,4$ from ABD, and  $a^{(\texttt{THO})}_{1,k}(t_l), \, k=1,\ldots,4$ and $\phi^{(\texttt{THO})}_{1,k}(t_l), \, k=1,\ldots,4$ from THO, where these estimates are denoted as $\hat a^{(\texttt{ABD})}_{1,k}(t_l)$, $\hat \phi^{(\texttt{ABD})}_{1,k}(t_l)$, $\hat a^{(\texttt{THO})}_{1,k}(t_l)$, and $\hat \phi^{(\texttt{THO})}_{1,k}(t_l)$.
These lead to harmonic representations of ABD and THO at time $t_l$:
\begin{align}
\bm{x}_{l,k}:=\big[\hat a^{(\texttt{ABD})}_{1,k}(t_l),\,\hat a^{(\texttt{THO})}_{1,k}(t_l),\,&\cos(2\pi\hat \phi^{(\texttt{ABD})}_{1,k}(t_l)),\,\cos(2\pi\hat \phi^{(\texttt{THO})}_{1,k}(t_l)),\nonumber\\
&\sin(2\pi\hat \phi^{(\texttt{ABD})}_{1,k}(t_l)),\,\sin(2\pi\hat \phi^{(\texttt{THO})}_{1,k}(t_l))\big]^\top\in \mathbb{R}^{6}.
\end{align}
Here, $\hat a^{(\texttt{ABD})}_{1,k}(t_l)$, $\cos(2\pi\hat \phi^{(\texttt{ABD})}_{1,k}(t_l))$ and $\sin(2\pi\hat \phi^{(\texttt{ABD})}_{1,k}(t_l))$ jointly represents the $k$-th harmonic of ABD. Note that sine and cosine functions are simultaneously taken into account since they fully capture the monotonically increasing phase function $\hat \phi^{(\texttt{ABD})}_{1,k}$. 
With ${\bm{x}}_{l}$, the Taken's lag map \citep{Takens:1981} is applied to integrate the temporal structure by
\[
\bm{x}_{l, \text{ lag}}:=[\bm{x}^\top_{l-9},\ldots,\bm{x}^\top_{l}]^\top\in \mathbb{R}^{240},
\] 
which will be used as the covariates for the GP regression.
Since in most cases the orders of magnitudes of amplitude and phase are different and their units are not comparable, $\bm{x}_{l, \text{ lag}}$ is normalized so that all components in the harmonic representation space are on the same scale \citep{cleveland1991local}. The use of normalization is needed here because the Euclidean distance in the harmonic representation space is used when fitting GP. To simplify the notation, the same notation $\bm{x}_{l, \text{ lag}}$ is used to denote the normalized $\bm{x}_{l, \text{ lag}}$.

With $\bm{x}_{l, \text{ lag}}$ and the concurrent airflow signal $y(t_{l})$, we are in the position to describe our \texttt{LocGP} fitting procedure.
First, the airflow signal is divided into non-overlapping time windows $\{I^{j}\}$, $j\in \mathbb{N}$. 
For a given window $I^j$ and a given time $t_{l_0} \in I^{j}$, to predict the airflow signal $g(t_{l_0})$,
the training data is constructed by taking the union of all the $K\in \mathbb{N}$ NNs in $I^{j}$ in the harmonic representation space, where those NNs are chosen from the {\em training pool} (i.e., the ``past'' time windows $\{I^{j'}\}$, where $j'<j$). Here, $K$ is a tuning parameter.\footnote{Note that there are other methods to define the local training set, and some discussions can be found at \cite{stein2004} and \cite{gramacy2015} in spatial and computer experiments context, respectively.} 
After obtaining the MLE of the GP for each time window, we predict the airflow signal using the plug-in estimate of the conditional mean at each time point.
Conditional standard deviation at each time point is used to quantify the pointwise prediction uncertainty.  To examine the improvement using GP, we also fit a locally stationary linear regression but without the inclusion of lag map (to avoid numerically instability due to the strong temporal dependence in SST features). We refer this benchmark model as \texttt{LocLm}.  

\begin{algorithm}[H] 
  \caption{\texttt{LocGP/LocDBGP} fitting procedure\label{algolocDBGP}}
  \begin{enumerate}
    \item[]
    \textbf{Inputs:} subject $s$ and the $j$-th time window $I^j$;
    $\{\bm{x}_{l, \text{ lag}}\}$, where $l \in I^{j}$ (i.e., the covariates in the testing window); $K$: \# of NNs; Prediction type. 
    
    \item If prediction $=$ \texttt{intra-subject}, search $K$-NNs from the training pool $\mathcal{F} = \{y_{l'}^{s}, \bm{x}_{l', \text{ lag}}^{s}\}_{l' \in \{I^{j'}\}}$, $j'<j$ where $\{y_{l'}^{s}, \bm{x}_{l', \text{ lag}}^{s}\}$ is the flow and harmonic representations pairs of the $s$-th subject; 
    
    \noindent if prediction $=$\texttt{inter-subject}, then the training pool is $\mathcal{F} = \bigcup_{s'} \{y_{l}^{s'}, \bm{x}_{l, \text{ lag}}^{s'}\}$, $s' \neq s$.
    
    \item Take union of the $K$-NNs of all harmonic representations within the $j$-th window to form the training set $\mathsf{T}_{j}$ and its corresponding response set $\mathsf{R}_{j}$.    

    \item Construct the covariance matrix $\bm{\Sigma}$ from $\mathsf{T}_{j}$. If $\texttt{DB}$ is considered, construct the covariance matrix as $D^{-1/2} \bm{\Sigma} D^{-1/2}$, where $D$ is a diagonal matrix with $D_{i,i} = \sum_{j} (\bm{\Sigma})_{i,j}$.
    
    \item Estimate GP parameters $(\mu, \sigma^{2}, \rho, \tau^{2})$ using MLE from $\mathsf{T}_{j}$ and $\mathsf{R}_{j}$. Denote the estimated parameters as $(\hat\mu, \hat\sigma^{2}, \hat\rho, \hat\tau^{2})$.

  \item[]
  \textbf{Outputs}:
  \begin{itemize}
      \item Point estimate: $\hat{\mu} + \hat{\mathsf{k}}[\hat{\bm{\Sigma}}+\hat{\tau}^2 I_{n}]^{-1}(\bm{y} -\hat{\mu} \bm{1})$.
      \item Standard deviation: $\text{diag(}\hat{\sigma}^2 - \hat{\mathsf{k}}[\hat{\bm{\Sigma}}+\hat{\tau}^2 I_{n}]^{-1}\hat{\mathsf{k}}^{\mathrm{T}})$.
  \end{itemize}
  
  \end{enumerate}
\end{algorithm}

\section{Material} \label{sec4}

\subsection{Databases}
The proposed method is examined by applying it to two datasets. The first dataset consists of 5 clinical subjects being evaluated for sleep apnea where their
whole night standard polysomnography recordings were used. The recordings were collected from the sleep center at Mackay Memorial Hospital (MMH), Taipei, Taiwan. The institutional review board of the MMH approved the study protocol (No. 18MMHIS142e). 
These 5 subjects were diagnosed to be free of sleep apnea (i.e., the apnea-hypopnea index is less than $5$) with recording length ranging from 6 to 7.5 hours. 
The THO and ABD were recorded by the piezo-electric bands and airflow was measured using nasal pressure, both at the sampling rate 100 Hz. All signals were acquired by a biosignal amplifier system from Embla (NeuroLite, Belp, Switzerland). These recordings will be integrated into the Taiwan Integrated Database for Intelligent Sleep (TIDIS) project, so we refer this database as {\em TIDIS}. 
To focus on evaluating our proposed model \texttt{LocGP} and to avoid the complicate signal quality issue commonly encountered in the biomedical signal, these recordings were visually screened and confirmed to be limited contaminated by artifacts.

The second database is from the Perelman School of Medicine, University of Pennsylvania (UPenn hereafter).  The UPenn dataset consists of a single subject used in \cite{mandel2016} with 829.3 seconds of 120 Hz signals of flow measured by spirometer and  ABD and THO signals by respiratory inductance plethysmography (RIP). The airflow prediction of the UPenn dataset represents a more challenge setting where we would need to predict the paradoxical breathing due to modulation of respiration by transitions from spontaneous to pressure support ventilation during general anesthesia.

\subsection{Evaluation}

We carry out both intra-subject and inter-subject predictions. In the intra-subject setup, we take the subject's historical data to train the prediction model, which will be used to predict the airflow signal in the future. In the inter-subject setup, we build the prediction model from a group of subjects, and apply the trained prediction model to predict the airflow signal of any subject not in that training group.

The following steps are common in both setups. First, the airflow, THO and ABD are resampled at $10$ Hz (i.e., $\Delta_t=1/10$). 
For STFT and SST, $h(t)=e^{-t^2/32}$ is chosen as the window. The frequency range of the SST is set to be from 0 to 2 Hz, and the frequency resolution is $10^{-4}$ Hz, and $b$, the frequency range is chosen to be $0.05$. For the curve extraction, $\lambda = 0.3$ is chosen.
Since the prediction is performed in a window-by-window fashion, the window size is set to be $30$ seconds, which is consistent with the epoch length commonly used in the sleep stage classification. For the intra-subject prediction, the first time window is excluded as some historical information is needed for the prediction. 
To determine the training data, the number of NNs at each time point is chosen to be $K=3$. The exponential covariance function is chosen when fitting \texttt{LocGP}.  

Define the root mean square error (RMSE) of the predicted airflow signal by $\|\hat{y}-y\|_{L^2(I^{j})}$, where $\hat{y}$ and $y$ are the predicted and true airflow signals respectively, and $I^{j}$ is the j\textsuperscript{th} window. 
Due to the inter-subject variability, there is an inevitable global phase shift of the predicted airflow signal and the true airflow signal when the inter-subject prediction is carried out. The global phase shift is then estimated and removed by a global alignment before carrying out our evaluation in the inter-subject prediction.
With the RMSE, over each window $I$, the {\em RMSE reduction} is defined as 
\[
1 - \frac{\|\hat{y}-y\|_{L^2(I)}}{\|y\|_{L^2(I)}}\,,
\]
which is our primary metric to assess the prediction performance. 
A large RMSE reduction indicates a good prediction performance. 

While the window-based RMSE reduction index is a desirable metric given the nonstationarity nature of the respiratory signal, it does not distinguish errors in different frequency bands. Since we want to recover detailed information encoded in airflow signal as accurately as possible, we shall further evaluate its performance in the high frequency region.   
To this end, the RMSE reduction of the differentiation of the predicted airflow signal, $\hat{y}'$, is evaluated against the RMSE reduction of the differentiation of the true airflow signal, $y'$ where a 6-th order Butterworth lowpass filter is applied to obtain these differentiated RMSEs. We call this index the {\em differentiation-RMSE reduction}.

In the following experiments, the computations were carried out using macOS Mojave 10.14.6, 3.2 GHz Intel Core i7, 32 GB 2667 MHz DDR4, with Matlab\textsuperscript{\textcopyright} version 9.8.0.1323502 (R2020a).

\section{Results} \label{sec5}

\begin{figure}[H]
    \centering
    \includegraphics[width=5.25in]{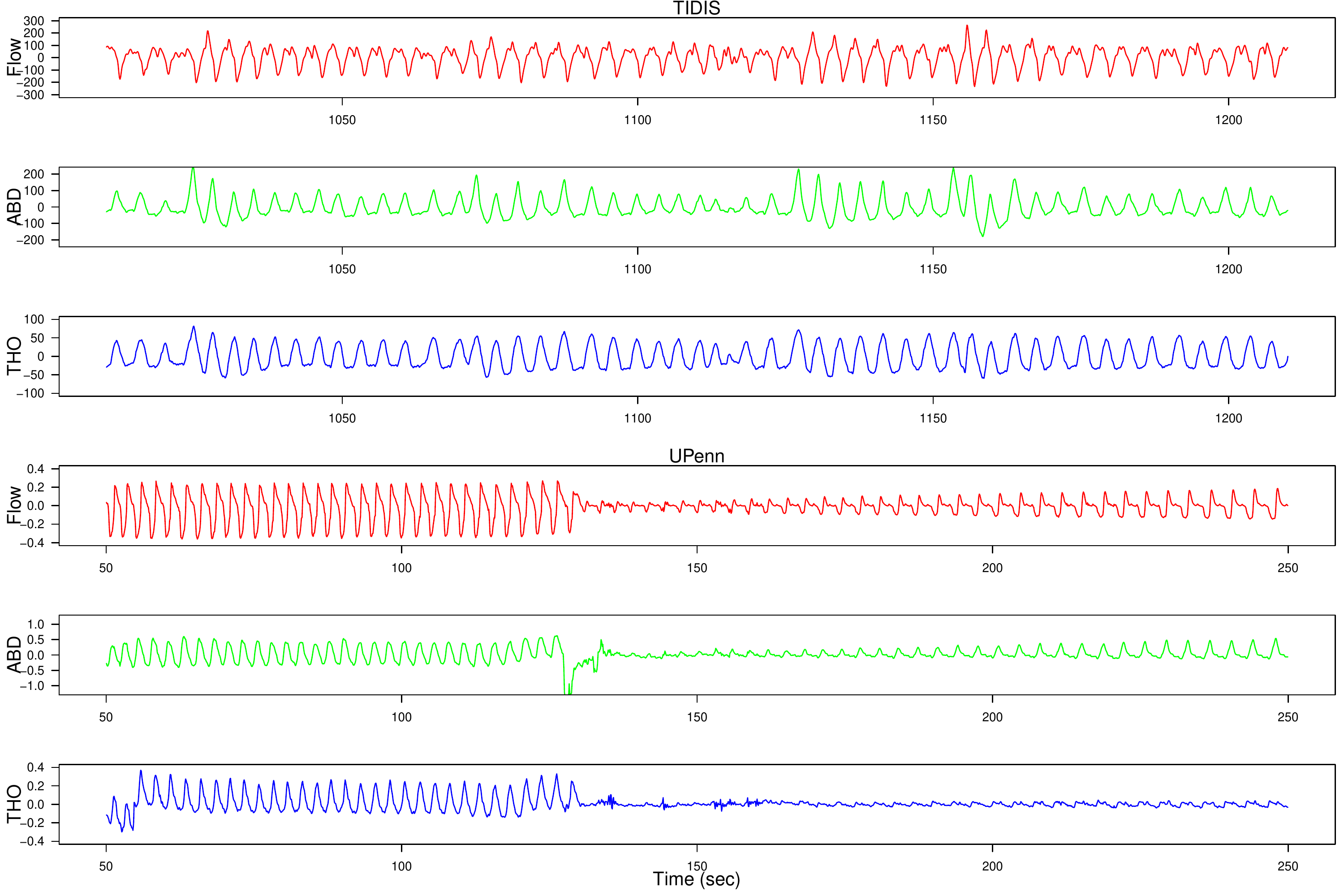}
    \caption{Detrended airflow (red), ABD (green), and THO (blue) signals used in the TIDIS and UPenn data analysis.}
    \label{fig:fig4}
\end{figure}

Fig.~\ref{fig:fig4} shows a segment of detrended signal of 200 seconds long from the TIDIS database (top three panels) and a segment of detrended signals with the same length from the Upenn database (bottom three panels). The airflow, ABD and THO signals all oscillate at the same rate, but with different oscillatory morphologies. Unlike the TIDIS database, we can clearly see a dramatic amplitude variation of the airflow, ABD and THO signals in the UPenn database.

\subsection{TIDIS}

\begin{figure}[htp!]
    \centering
    \includegraphics[width=5.25in]{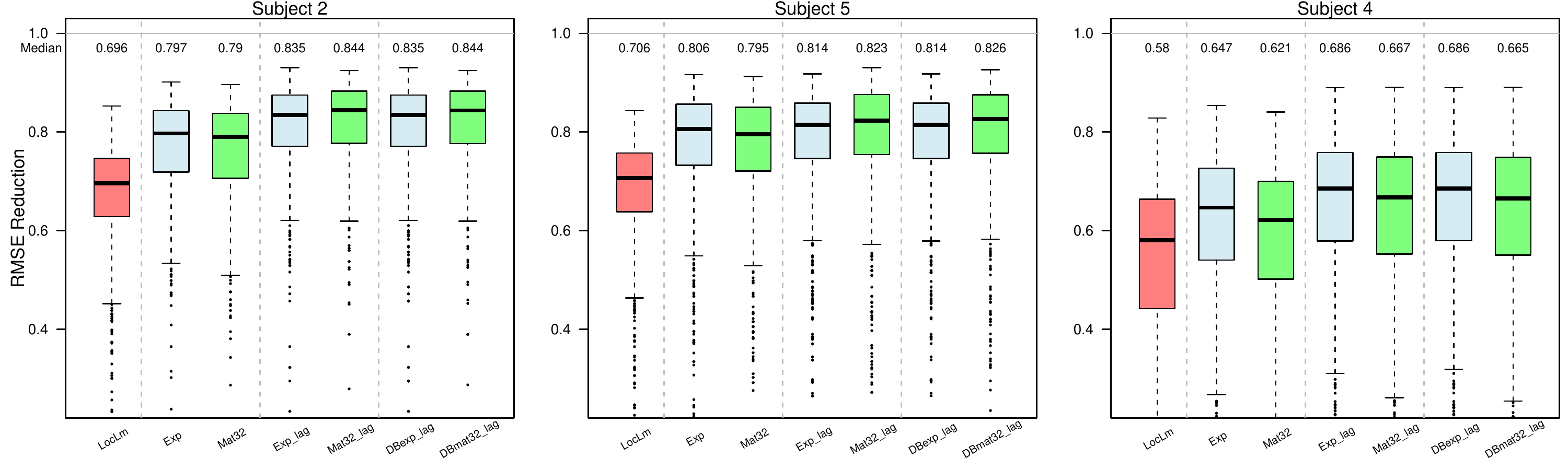}\\
    \caption{Boxplots of the intra-subject RMSE reduction. The subjects shown here represent the best (Subject 2), average (Subject 5), and the worst (Subject 4) predictions. Red boxplots are for \texttt{LocLm}. The second sub-panel within each panel includes the cases that the normalized SST features were used, the third sub-panel includes the cases where both normalization and lag map have been used, and \texttt{LocDBGP} with normalization and lag map are shown in the fourth sub-panel.  Lightblue and green boxplots are results with the exponential and Mat\'ern $\nu=1.5$ covariances.}
    \label{fig:TIDIS_RMSE_boxplots}
\end{figure}

We first summarize the results for the intra-subject prediction. In Fig.~\ref{fig:TIDIS_RMSE_boxplots}, the boxplots of RMSE reduction for the intra-subject prediction over all windows of those subjects with the best, average and worst prediction performance are shown. Overall, the \texttt{LocGP} and \texttt{LocDBGP} with the Mat\'ern covariance function $\nu = 1.5$, the lag map, performs the best (except for Subject 4). However, this choice leads to less well calibrated prediction uncertainty. Therefore, we will report the results using the exponential covariance function.  
It is also clear from the boxplot that a good amount of improvement is achieved compared with \texttt{LocLm}. 
The median differentiation-RMSE reduction for \texttt{LocGP} and \texttt{LocDBGP} are nearly identical, and those reductions are 0.732, 0.750, 0.768, 0.569, 0.743 for Subject 1 through Subject 5, respectively.
The median empirical coverage rate of the point-wise 95\% confidence interval across all the time windows are also reported. The median coverage rates of \texttt{LocGP} (\texttt{LocDBGP}) are 0.930 (0.883), 0.933 (0.887), 0.937 (0.900), 0.903 (0.850), 0.937 (0.907) for Subject 1 through Subject 5, respectively. The coverage rates of \texttt{LocGP} are closer to the nominal rate than that of the \texttt{LocDBGP}. 
Nonetheless, both \texttt{LocGP} \texttt{LocDBGP} yield reasonable prediction uncertainty quantification.

\begin{figure}[htp!]
    \centering
    \includegraphics[width=4.5in]{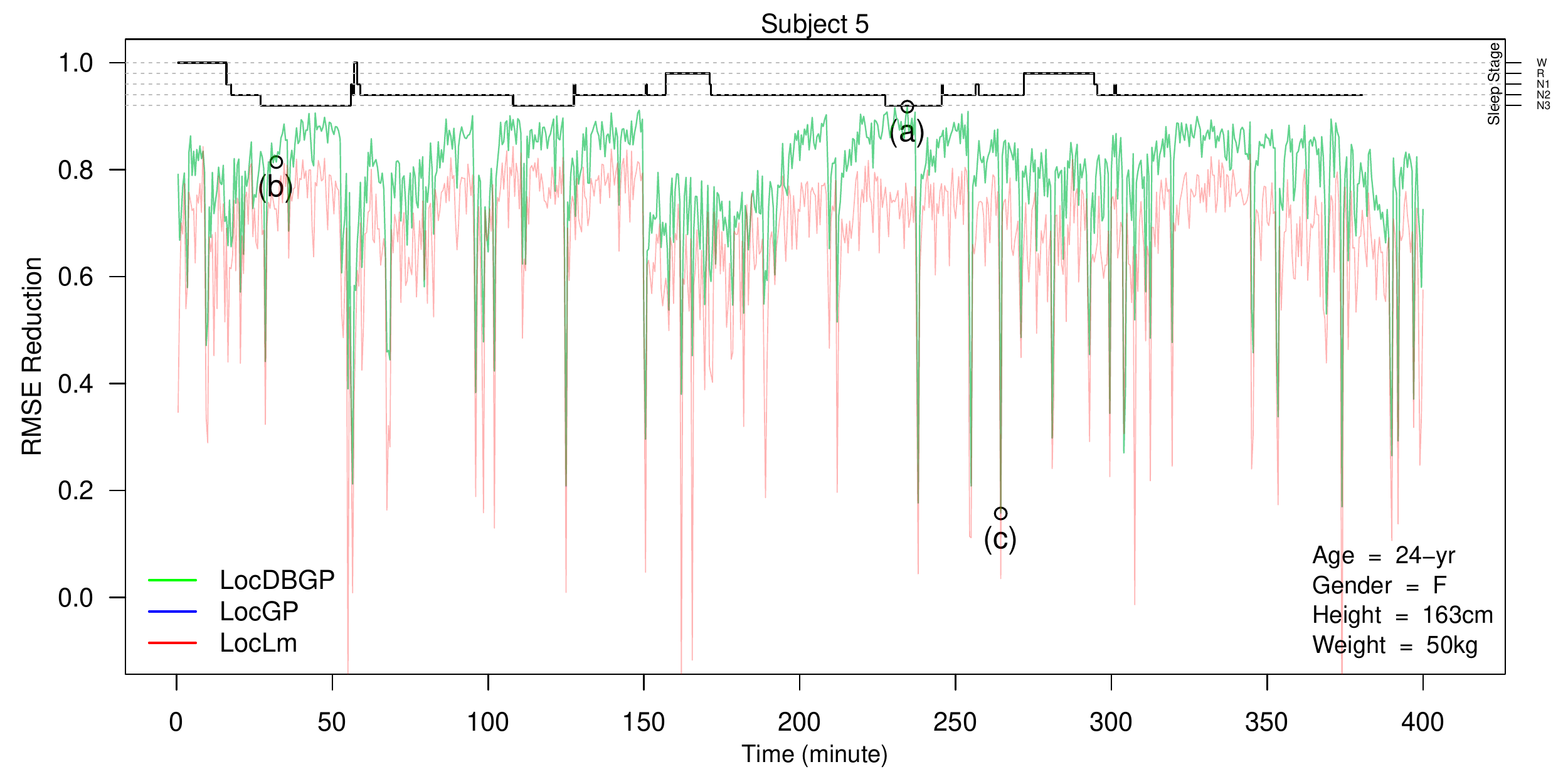}
    \includegraphics[width=5.5in]{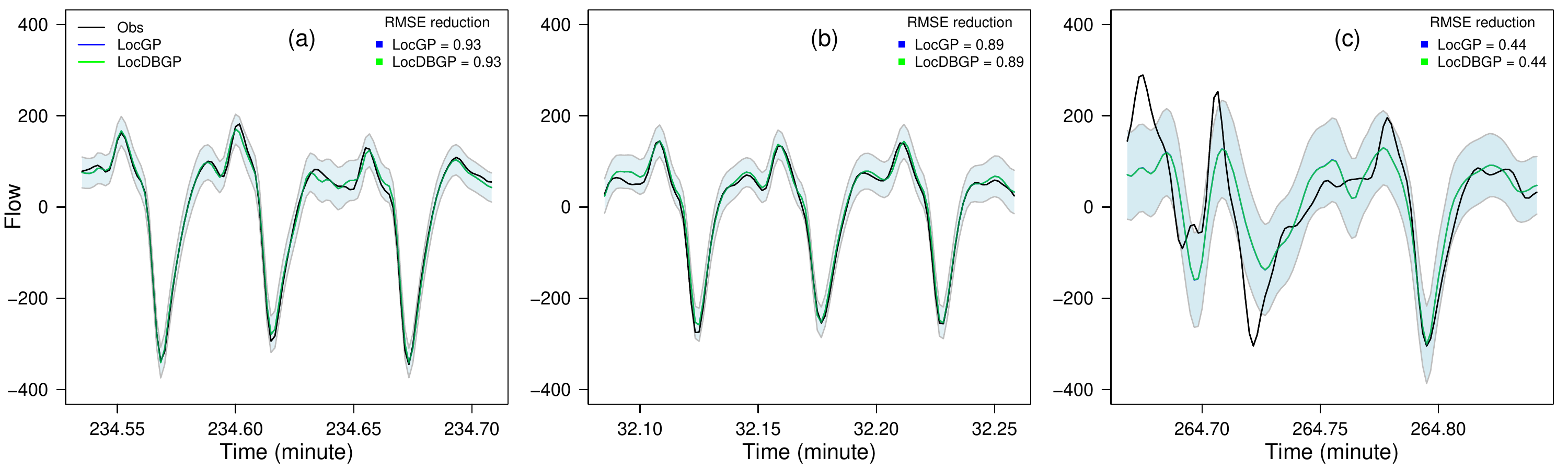}

    \caption{RMSE reductions over consecutive 30-second windows for Subject 5 within-subject prediction is shown on the top panel. The sleep stage is included for a comparison. Since the RMSE reductions between \texttt{LocDBGP} and \texttt{LocGP} are almost identical, only the RMSE reductions for \texttt{LocDBGP} (green) and \texttt{LocLm} (red) are shown here. In the bottom panel, we show examples of ``best'' (a), ``average'' (b), and a ``poor'' prediction (c). To enhance the visualization, we only show 10 seconds segment. The envelope in each case represents the \texttt{LocGP} pointwise 95\% confident interval.}
    \label{fig:TIDIS_IP}
\end{figure}

Next, we focus on Subject 5, which represents the ``average'' case among all 5 subjects in the TIDIS dataset. In the top panel of Fig.~\ref{fig:TIDIS_IP}, the RMSE reduction over consecutive 30-second windows is shown. The predictions are not accurate over the first few windows due to the limited historical signals. However, the predictive power improves as the observational history increases.

Physiologically, the prediction performance may depend on different sleep stages.  According to American Academy of Sleep Medicine \citep{berry2012aasm}, there are five sleep stages: wake (W), rapid-eye-movement (R), and non rapid-eye-movement that can further divided into N1, N2, and N3. See the top panel of Fig.~\ref{fig:TIDIS_IP} for an example of the whole night sleep dynamics. It is a physiological fact that the respiratory activity during N2 and N3 is more stable compared with that during W, R and N1. Therefore, we would expect a higher RMSE reduction during N2 and N3. To confirm this hypothesis, we divide all 30 seconds windows into two groups. The first group contains those windows that have the same sleep stages as their following windows, and the second group contains those windows that have different sleep stages as their following windows. In other words, the second group contains windows that have sleep stage jumps.
The median of overall RMSE reduction of all windows of 5 subjects with (without) sleep stage jumps are 0.845 (0.885), 0.789 (0.834), 0.677 (0.696), 0.745 (0.766), 0.724 (0.754) for N3, N2, N1, REM and W. As expected, when there are changes in the sleep stage, the RMSE reduction is lower, and the RMSE reductions are usually high when the sleep stage is N2 or N3.

In order to illustrate the quality of predictions, three examples that represent the ``best'', ``average'' and ``poor'' cases of Subject 5 are plotted in the bottom panel of Fig.~\ref{fig:TIDIS_IP}. Here, it is clear that \texttt{LocDBGP} and \texttt{LocGP} both predict the airflow well and their fits are nearly identical. The ``poor'' case shown here presents a typical example of how the artifacts deteriorate the prediction performance.

\begin{figure}[hbp!]
    \centering
    \includegraphics[width=5.5in]{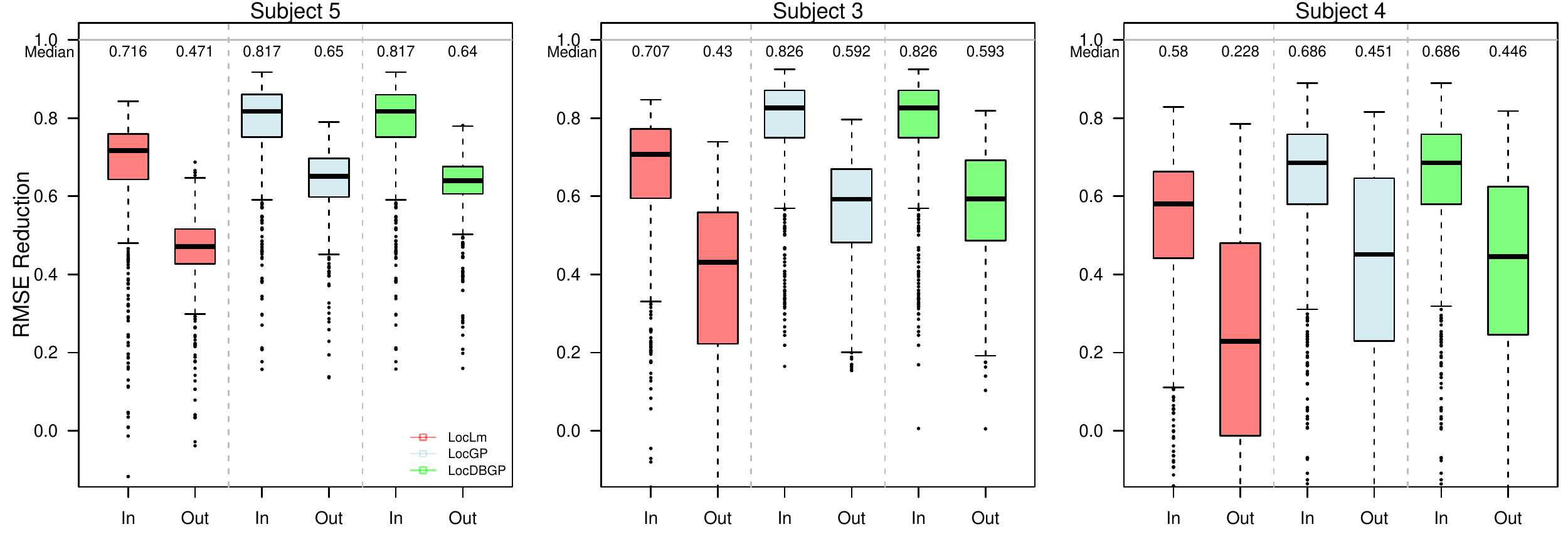}\\
    \caption{Boxplots of the RMSE reduction of all windows for intra-subject and inter-subject prediction. The ``best'', ``average'' and ``worst'' ones, Subjects 5, 3 and 4, are shown on the left, middle and right panels respectively. In each sub-panel, the intra- and inter- subject results are shown on the left and right respectively.  
    As in Fig.~\ref{fig:TIDIS_RMSE_boxplots}, red boxplots are for \texttt{LocLm}, lightblue boxplots are for Mat\'ern $\nu=1.5$ covariance and green boxplots are for exponential covariance.}
    \label{fig:TIDIS_RMSE_Out_boxplots}
\end{figure}

Next, the performance of the inter-subject prediction is evaluated. Here, the signals from Subjects 1 and 2 are used to establish the prediction model for the flow signals, and the model is applied to Subjects 3, 4, and 5.
The global alignment to correct the global phase shift are 0.1 second forward, 0.1 second and 1 second backward for Subjects 3, 4 and 5, respectively.
Fig.~\ref{fig:TIDIS_RMSE_Out_boxplots} shows the boxplots of the RMSE reduction over all windows. We also add the corresponding intra-subject prediction for a comparison to contrast the inter-subject and intra-subject predictions. The median differentiation-RMSE reduction for \texttt{LocGP} (\texttt{LocDBGP}) are 0.451 (0.453), 0.294 (0.288), 0.563 (0.561).
It is not surprising that the performance of inter-subject prediction is worse than those of the intra-subject due to the inter-subject variability. However, the predictions are still reasonable with a substantial improvements compare with the \texttt{LocLm} predictions. 
For different sleep stages, the median of RMSE reductions of those windows of 5 subjects with (without) sleep stage jumps are 0.581 (0.668), 0.597 (0.639), 0.534 (0.553), 0.455 (0.455), 0.574 (0.651) for N3, N2, N1, REM, W, respectively. It is also not surprising that the empirical coverage rate decreases for the inter-subject prediction here. The median of the coverage rates of \texttt{LocGP} (\texttt{LocDBGP}) are 0.717 (0.687), 0.700 (0.640), and 0.650 (0.583) for Subjects 3, 4 and 5.

\begin{figure}[htp!]
    \centering
    \includegraphics[width=4.5in]{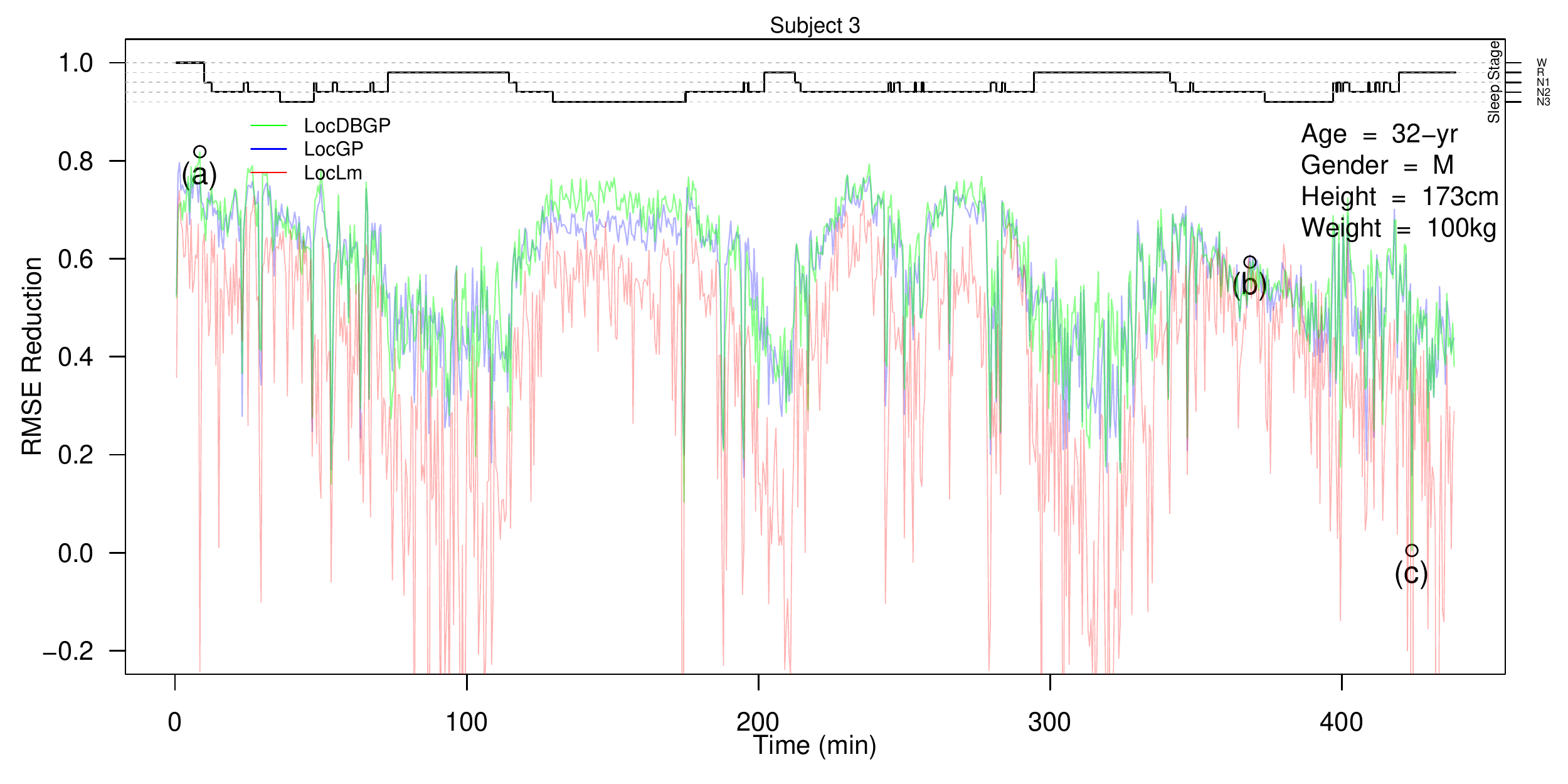}
    \includegraphics[width=5.5in]{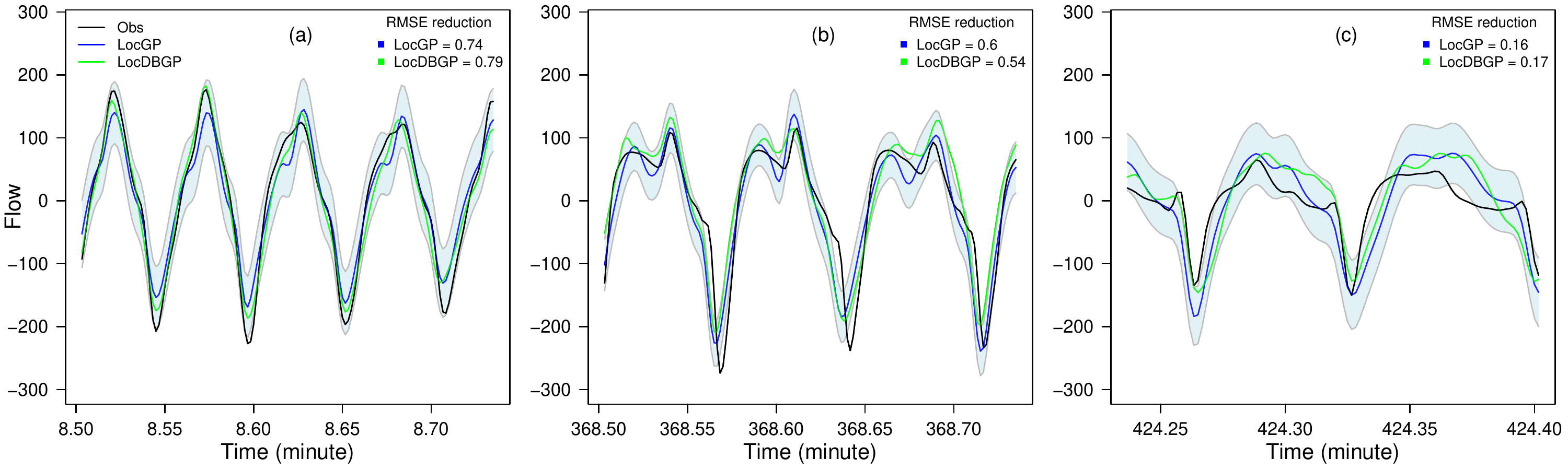}
    \caption{As in Fig.~\ref{fig:TIDIS_IP} but for the inter-subject prediction of Subject 3.}
    \label{fig:TIDIS_OP}
\end{figure}

As in the assessment of intra-subject prediction, we examine the ``average'' case, the Subject 3 shown in Fig.~\ref{fig:TIDIS_OP}. In the top subplot of Fig.~\ref{fig:TIDIS_OP}, it is clear that compared with the intra-subject prediction, the prediction in the first few windows are reasonably well. It is because in the inter-subject prediction, we do not need to accumulate historical data before establishing a reasonably well prediction model. 

Lastly, we zoom in several 10-second time windows in Subject 3 to both qualitatively and quantitatively assess the inter-subject predictions. In the bottom subplot of Fig.~\ref{fig:TIDIS_OP}, three windows associated with the ``best'', ``average'', and ``poor'' prediction performances are shown. Despite the less impressive numerical performance in terms of RMSE reduction, the ``best'' and the ``average'' cases capture the flow signals reasonably well, and the overall oscillatory morphology is well captured. The ``poor'' case here again demonstrates that, as would be expected, the prediction performance degrades when the respiratory signals are contaminated by artifacts.

\subsection{UPenn}

\begin{figure}[htp!]
    \centering
    \includegraphics[width=4.5in]{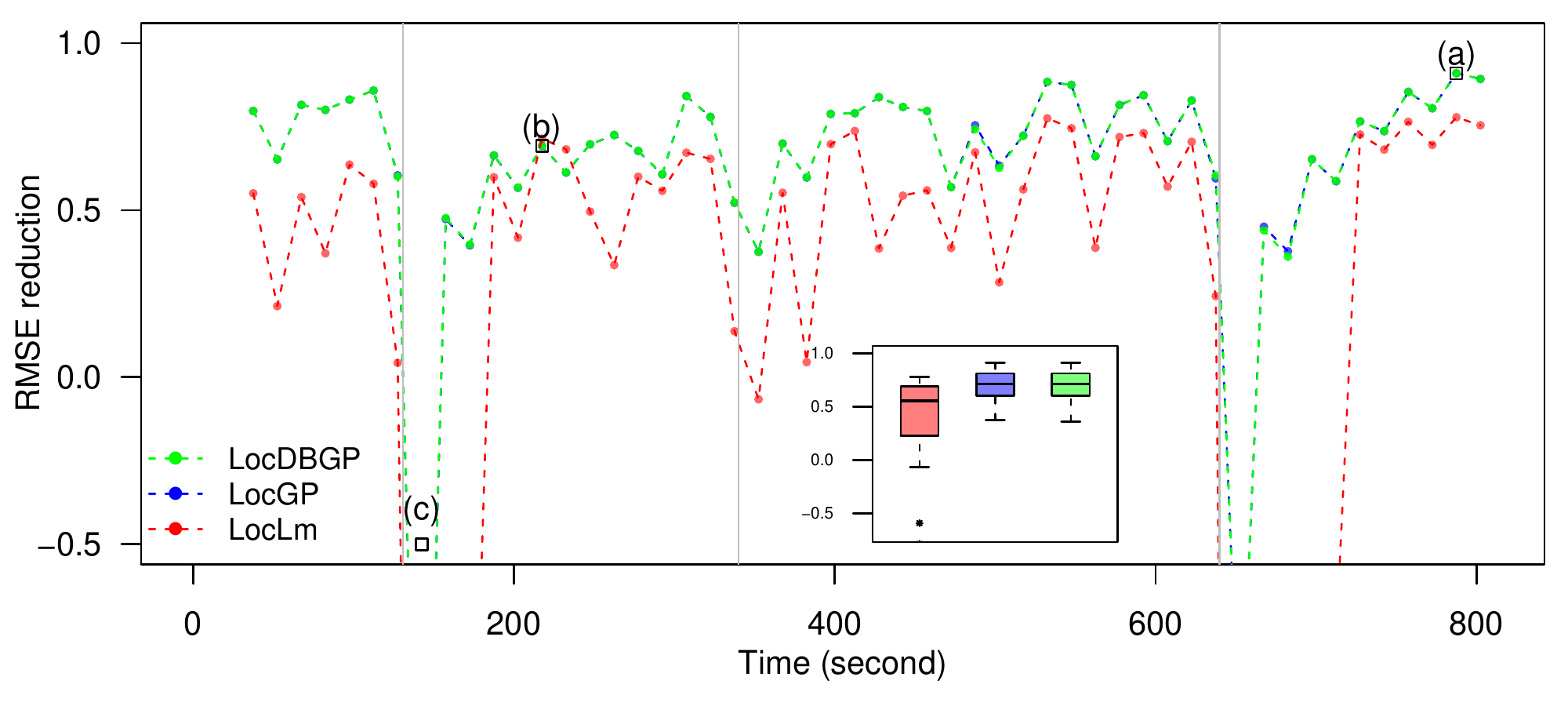}
    \includegraphics[width=5.5in]{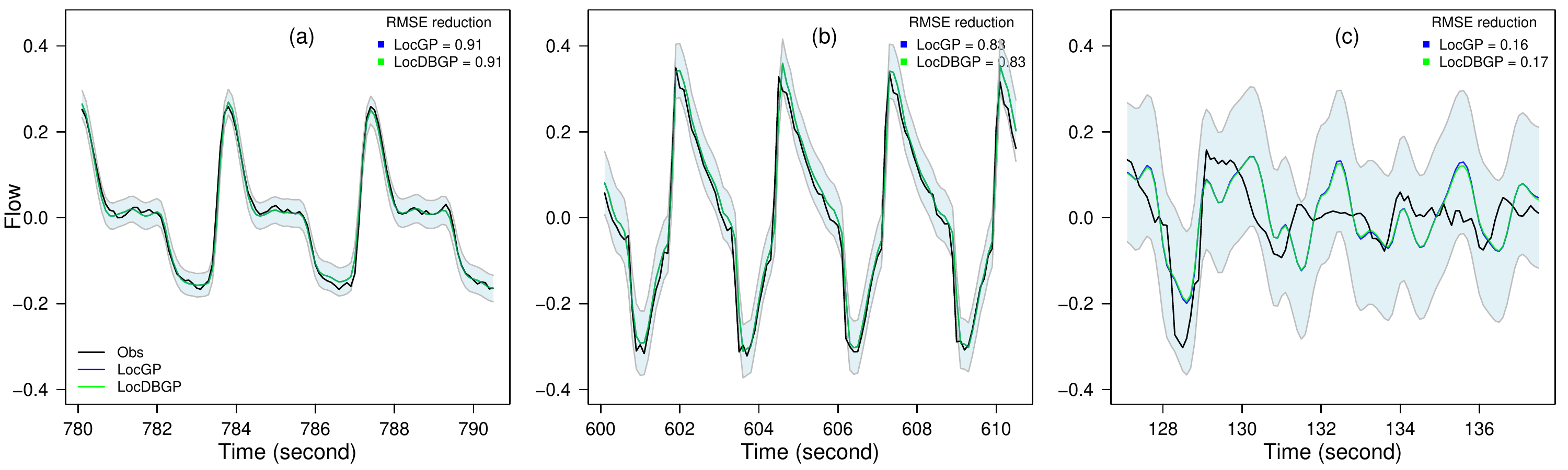}
    \caption{
    The RMSE reduction over consecutive windows for the Upenn within-subject prediction is shown on the top panel.
        Here the RMSE reductions are calculated every 15 seconds for \texttt{LocDBGP} (green), \texttt{LocGP} (blue), and \texttt{LocLm} (red). The vertical gray lines are those time points when transitions have occurred. 
        Boxplots of the RMSE reduction (across all the 15-second windows) are plotted in the subplot. 
    In the bottom panel, we show examples of ``best'' (left), ``average'' (middle), and a ``poor'' prediction. To enhance the visualization, we only show 10 seconds segment. The envelope in each case represents the pointwise 95\% confident interval.
    The poor prediction shown here occurred during the first transition time period.}
    \label{fig:JeffRMSE_reduce_ts}
\end{figure}

The intra-subject prediction for the UPenn data is demonstrated. This database represents a more challenging paradigm in that we need to adapt to the 
paradoxical breathing due to modulation of respiration by transitions from spontaneous to pressure support ventilation during general anesthesia.
The top subfigure of Fig.~\ref{fig:JeffRMSE_reduce_ts} shows the RMSE reduction of the prediction using \texttt{LocGP} and \texttt{LocDBGP}, respectively. Both fits are based on the exponential covariance function, normalized SST with lag map. We observe that both \texttt{LocDBGP} and \texttt{LocGP} preform reasonably well, except for those windows when dramatic changes in the oscillatory pattern occur (i.e, around these vertical gray lines, and in the middle of the third segment). 
The results suggest that
the GP model tends to be more robust against sudden change of signals compared with linear regression model as these drops in RMSE reduction are smaller. The median coverage rates of \texttt{LocGP} and \texttt{LocDBGP} are 0.950 and 0.825 and the median differentiation-RMSE reduction are both 0.6794.         

A closer look into how the \texttt{LocGP} and \texttt{LocDBGP} fit during three ``representative'' windows is shown in the bottom subfigures of Fig.~\ref{fig:JeffRMSE_reduce_ts}. The left panel represents the ``best'' case, where the algorithm performs well, especially \texttt{LocDBGP}, with the RMSE reduction close to 1. The middle panel shows an ``average'' case, where \texttt{LocDBGP} captures the general flow pattern while \texttt{LocGP} somewhat underestimate the amplitude. The right panel gives an example of a ``poor'' case, where both \texttt{LocGP} and \texttt{LocDBGP} got confused when there exists a sudden change of the oscillatory pattern. 

In the supplementary materials, we provide more numerical results, including results with different covariance functions, different choice of $K$, skipping the diffusion, and skipping the normalization.

\section{Discussion and Conclusion} \label{sec6}

In this work, the \texttt{LocGP} and \texttt{LocDBGP} models are proposed to predict the airflow signal from the simultaneously recorded ABD and THO signals. This problem is challenged due to the need to handling the complicated non-stationary oscillation of the respiratory signal, and the nonlinear relationship among those signals. 
The main novelty of this work is twofold. First, we propose a novel harmonic representation of the respiratory signals by applying SST. Second, we develop a locally stationary GP, \texttt{LocGP}, to model the nonlinear relationship among the airflow, ABD and THO signals. The \texttt{LocGP} is enriched by capturing the geometric structure by the diffusion idea, which leads to \texttt{LocDBGP}. The \texttt{LocGP} not only enables the computational scalability but also provides a means to handle different types of non-stationarity. We evaluate our method in both the intra-subject and inter-subject setups and obtain encouraging results.  To our knowledge, this is the first algorithm able to handle such a challenge.

Compared with the EMD algorithm considered in \cite{mandel2016}, the new harmonic representation is both theoretically well-founded and interpretable. We expect that this representation has a potential to further evaluate the respiratory dynamics for patients with different diseases.
The proposed model may also be useful in estimating other physiologic signals such as arterial blood pressure, from nonintrusive surface sensors.
%

The study has several limitations.
First, the number of components to use in SST is arbitrarily chosen as four, as this setting provides good reconstructions of ABD and THO. It would be interesting to investigate how to determine the optimal number of components. The curve extraction algorithm \eqref{CurveExtractionFormula} is critical for constructing the harmonic representation. While it works reasonably well in this work, we notice that it is also a source of error. Specifically, a fixed $\lambda$ that works well for some segments of recordings might not work that well for other segments. We will explore the solution in our upcoming clinical work.
Second, we only consider the simple covariance structure with the Euclidean distance of the standardized harmonic representation when fitting GP. In general, other forms of covariance function can be used; for example, the product form or additive form. These general forms might have their own benefits, particularly when the amplitude and frequency are of different importance. How to construct a more sensible covariance function is another future direction. 
Third, the inter-subject variability is not handled in this study. We simply take the first two subjects (according to the chronological order) to construct the prediction model and apply it to the remaining subjects. It would be beneficial to take the inter-subject variability into account when constructing a prediction model that can more sensibly and effectively ``borrow strength'' across individuals to achieve a better inter-subject prediction. 
Fourth, the recordings in the TIDIS database are confirmed visually to be relatively less impacted by artifacts. Since artifacts would inevitably reduce the prediction performance, to apply the established algorithm to the clinical setup, establishing a proper signal quality index so that we can properly identity respiratory signal segments of poor quality is needed. To avoid distracting the focus of this paper, we postpone a development of such a robust procedure to our future work. 
Last but not the least, it should be emphasized that the breathing pattern can be controlled voluntarily. For example, when we are talking, the flow signal might deviate significantly. When a subject is under some pathological situations, like hemothorax or pneumothorax, the relationship between the rib cage volume and lung volume is different, and the above-mentioned physiological relationship might not hold. In this work, these voluntary activities and pathological situations are not discussed. These important topics will be explored in our future work.

\section*{Acknowledgements}
This work is part of the Taiwan Integrated Database for Intelligent Sleep (TIDIS) project support by Ministry of Science and Technology 109-2119-M-002-014-, NCTS Taiwan.

\bibliographystyle{plain} 
\bibliography{BTDA.bib}       %

\clearpage
\appendix

\section{SST reconstruction}

\begin{figure}[bhp!]
    \centering
    \includegraphics[width=5.5in]{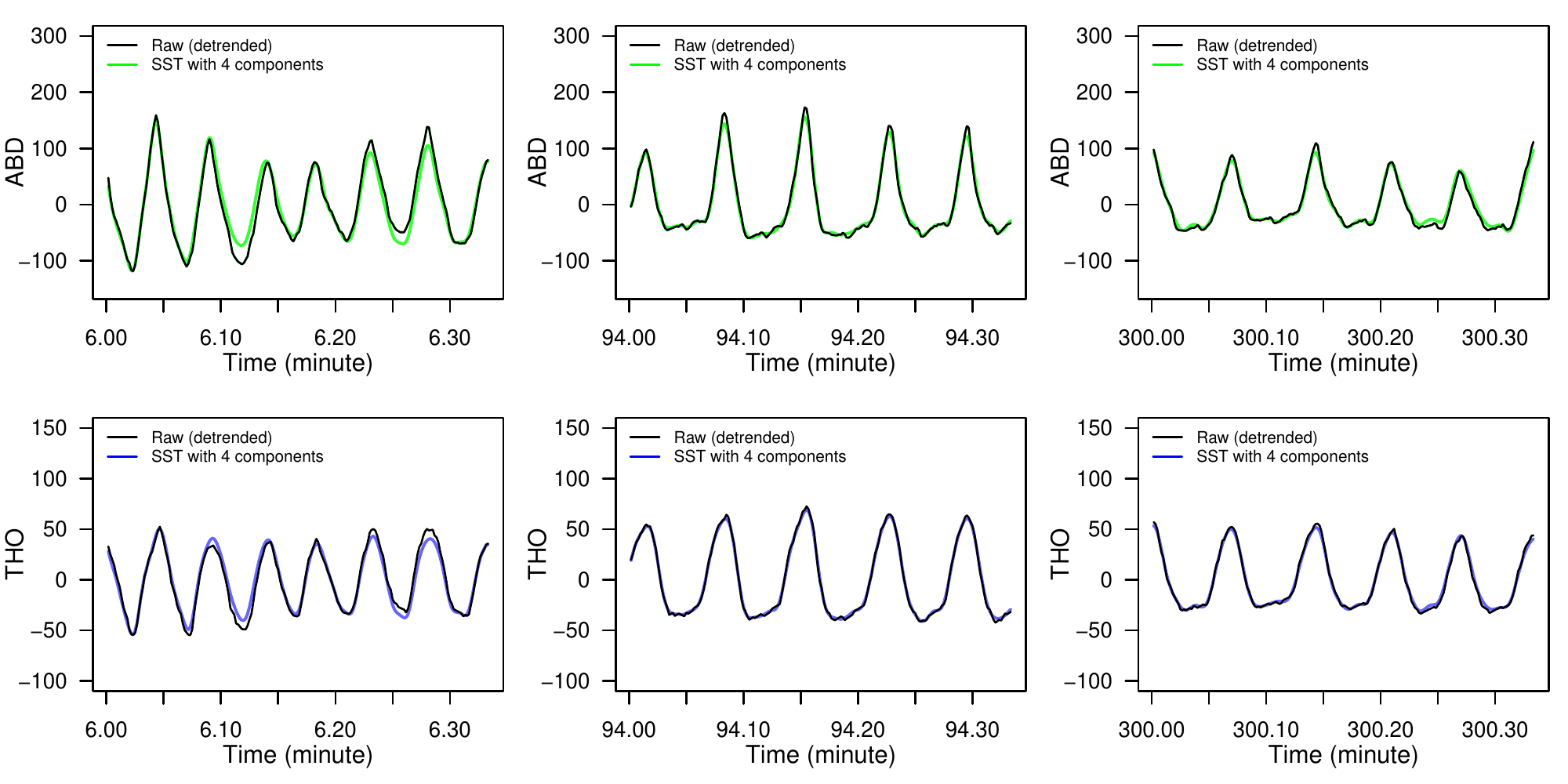}
    \caption{The reconstructed ABD (green) and THO (blue) signals of three segments from Subject 2 in the TIDIS dataset. The reconstruction is from the first 4 harmonics decomposed by SST.}
    \label{fig:figS0}
\end{figure}

We show several examples in Fig.~\ref{fig:figS0} to demonstrate that SST can efficiently decompose ABD and THO into their harmonics. 

\section{The choice of covariates: SST versus time series}
\begin{figure}[H]
    \centering
    \includegraphics[width=4.5in]{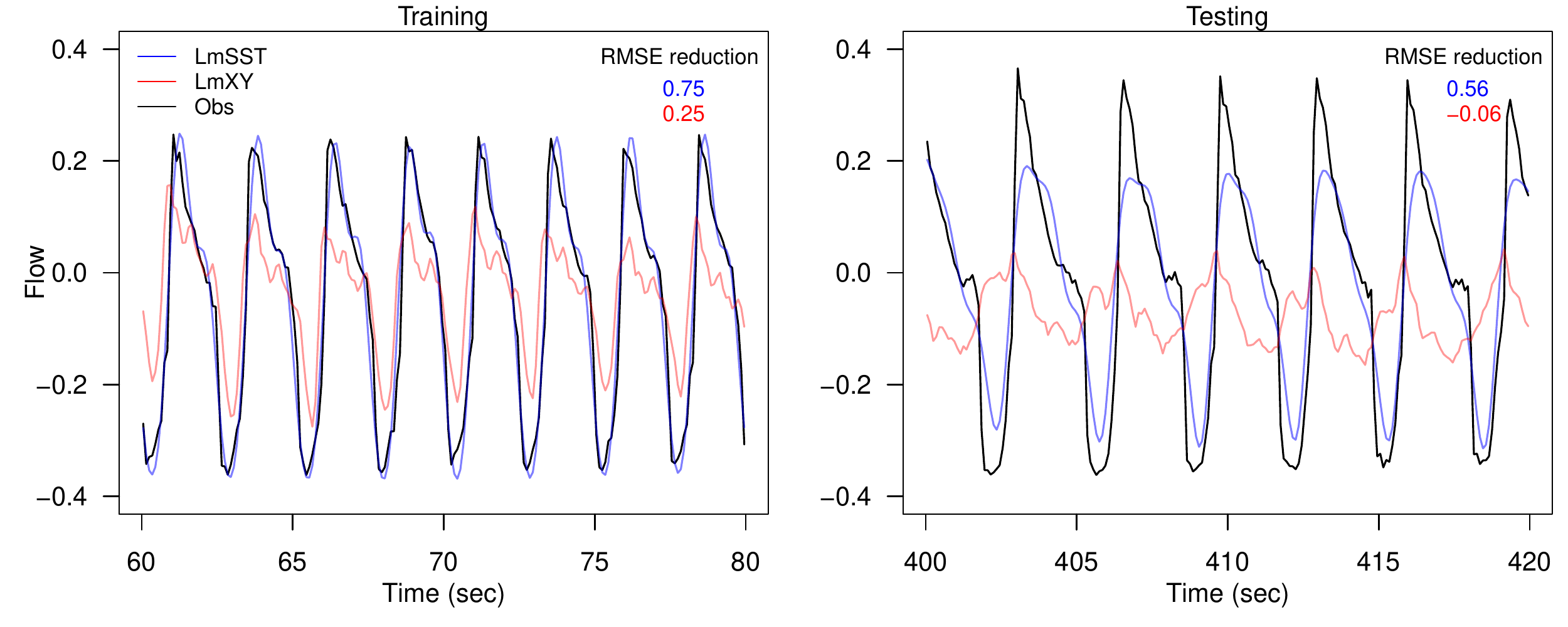}
    \caption{Airflow predictions of the UPenn data for training and testing periods using linear regression with harmonic representation features and time domain features as covariates, respectively.}
    \label{fig:figS1}
\end{figure}

An example is provided to demonstrate the prediction improvement by using harmonic representation features rather than that of the time series features.
Fig.~\ref{fig:figS1} shows in-sample (left panel) and out-of-sample (right panel) predictions for linear regression using harmonic representation features as covariates (blue lines, \texttt{LmSST}, which is \texttt{LocLm} in the main article) and time series as covariates (pink lines, \texttt{LmXY}). The main message here is that, linear regression with harmonic representation features can often achieve a fairly reasonable prediction performance and the improvement compared with the time-domain linear regression is usually substantial.            

\begin{figure}[H]
    \centering
    \includegraphics[width=4.5in]{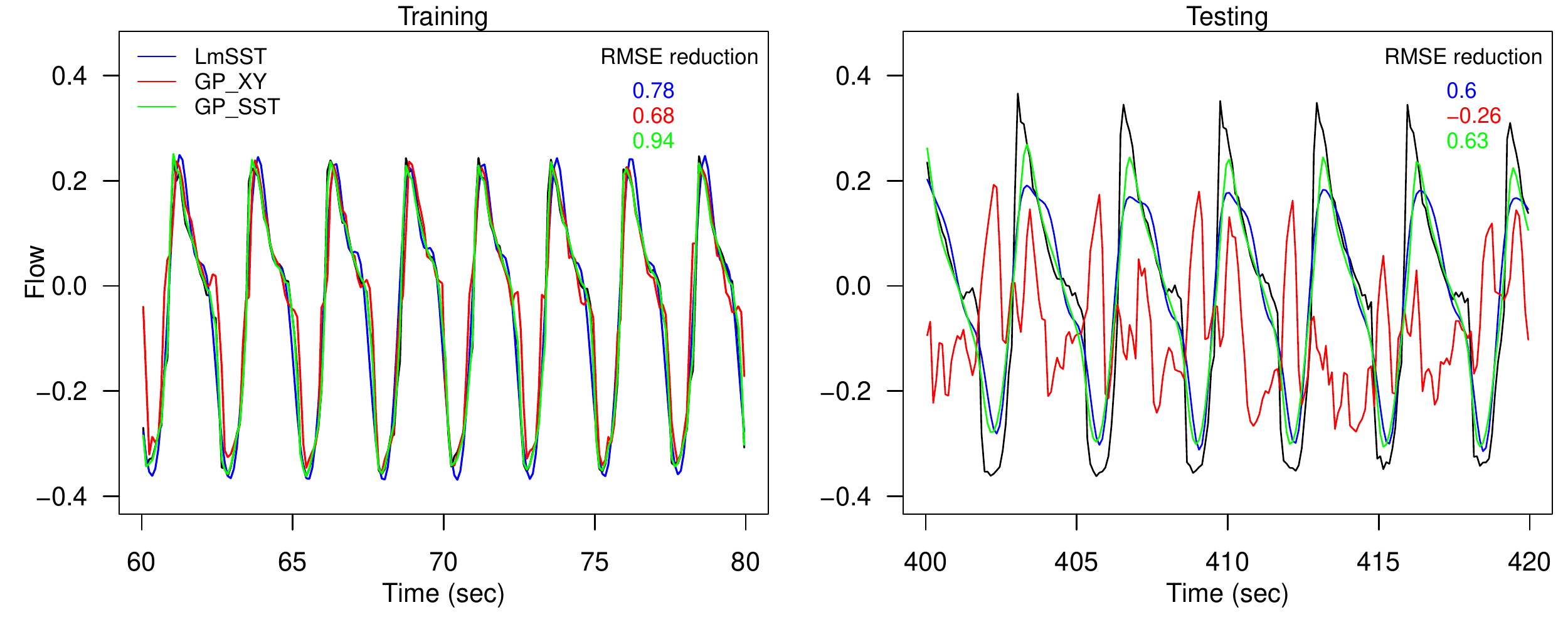}
    \caption{As in Fig.~\ref{fig:figS1}, but here the comparison of predictions using linear regression with harmonic representation as covariates (blue), GP with time domain features as covariates (red), and GP with harmonic representation as covariates (green), respectively, are shown.}
    \label{fig:figS2}
\end{figure}

In Fig.~\ref{fig:figS2}, the prediction results with the same training and testing sets but with Gaussian process (GP) regression with exponential covariance function are shown. Here, the GP fitting is carried out using both harmonic representation features (green) and time-domain features (red). A couple of observations follow: 1) GP usually achieves a good in-sample prediction, especially with harmonic representation feature; 2) the GP out-of-sample prediction using time-domain features as covariates can preform substantially worse than linear regression with harmonic representation features (blue).        

An important conclusion here is that using harmonic representation features to represent the oscillatory signals to conduct regression analysis usually leads a good prediction performance, even with a simple model such as linear regression. A combination of SST and GP further improves the prediction accuracy.

\section{Standardization of SST features}

\begin{figure}[H]
    \centering
    \includegraphics[width=5in]{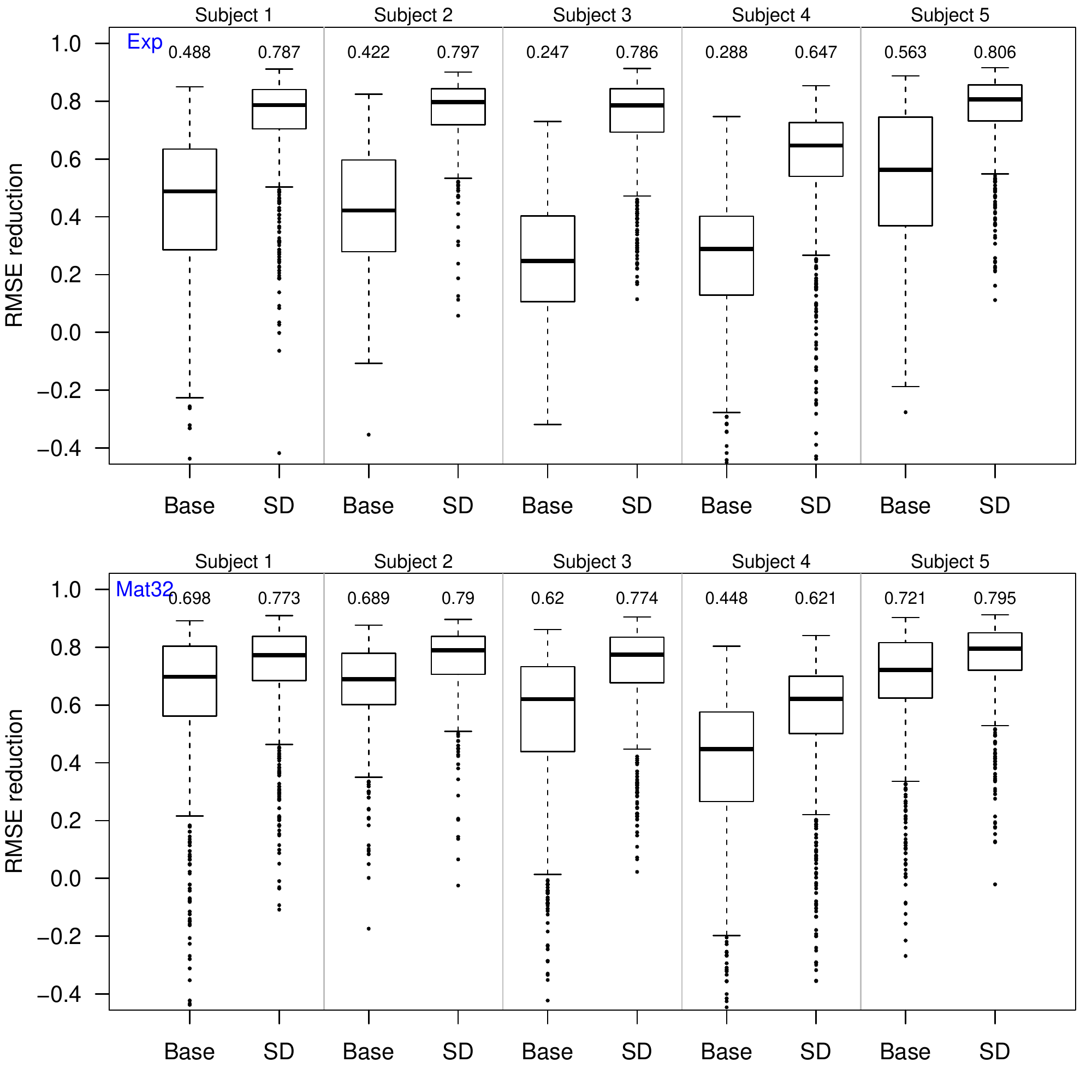}
    \caption{RMSE reduction of each subject with (SD) and without (Base) standardization of the harmonic representation features when fitting GP using exponential (Exp) covariance (upper) and Mat\'ern $\nu = 1.5$ (Mat32) covariance (lower), respectively.}
    \label{fig:figS3}
\end{figure}

The effect of normalization in terms of prediction by standardizing the harmonic representation space is examined here. After the standardization of the harmonic representation space, all components of the harmonic representation are in the same scale. Note that the scale of amplitude components are usually different from that of phase components, which are always between -1 and 1. In Fig.~\ref{fig:figS3}, the results from the TIDIS dataset confirms that standardization is needed.       

\begin{figure}[H]
    \centering
    \includegraphics[width=5in]{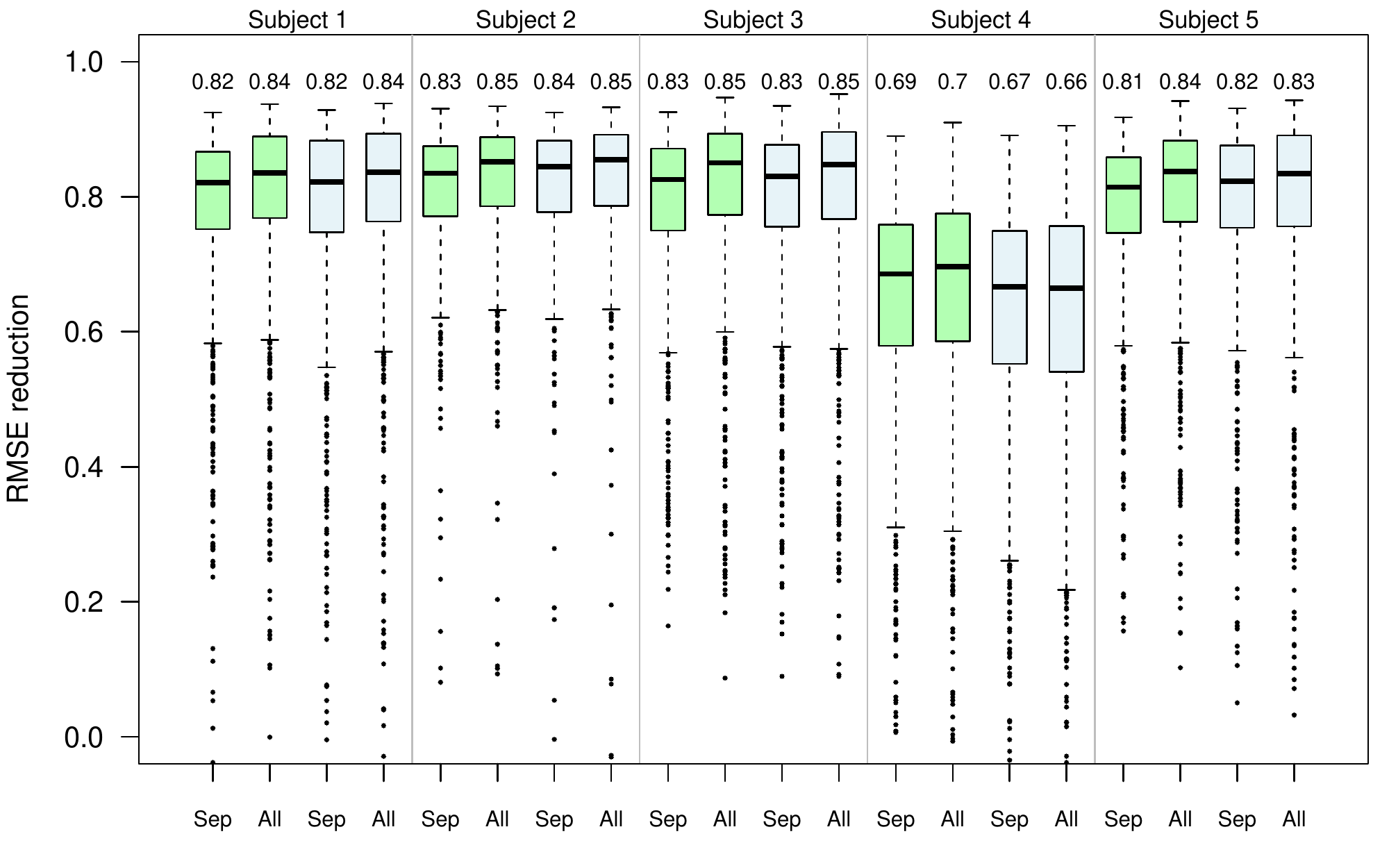}
    \caption{Bxoplots of RMSE reduction when applying standardization separately (``Sep'') and all-together (``All'') for exponential (green) and Mat\'ern $\nu = 1.5$ (lightblue) covariances.}
    \label{fig:figS4}
\end{figure}

In the main text of the manuscript, we standardize the harmonic representation features for training and testing datasets \textit{separately}. In Fig.~\ref{fig:figS4}, it is shown that the prediction performance further improves if we apply a single standardization for the whole data, including both training and testing datasets. Note that the standardization for training and testing datasets separately is what we can carry out in the real world application.

\section{The choice of covariance function}

A summary of prediction performance using different covariance functions when fitting \texttt{LocGP} is explored here. The results suggest that GP with the Mat\'ern $\nu = 1.5$ covariance trends to perform slightly better than the exponential and the squared exponential covariances in terms of RMSE reduction. However the asscoiated prediction uncertainty, evaluated by empirical coverage rate, tends to be the worst (a bit less than 80\% coverage rate for its 95\% confident interval). It is also worth pointing out that, despite commonly used when performing non-parametric regression with GP, the squared exponential performs the worst in all the cases we have examined.      

\begin{figure}[H]
    \centering
    \includegraphics[width=5in]{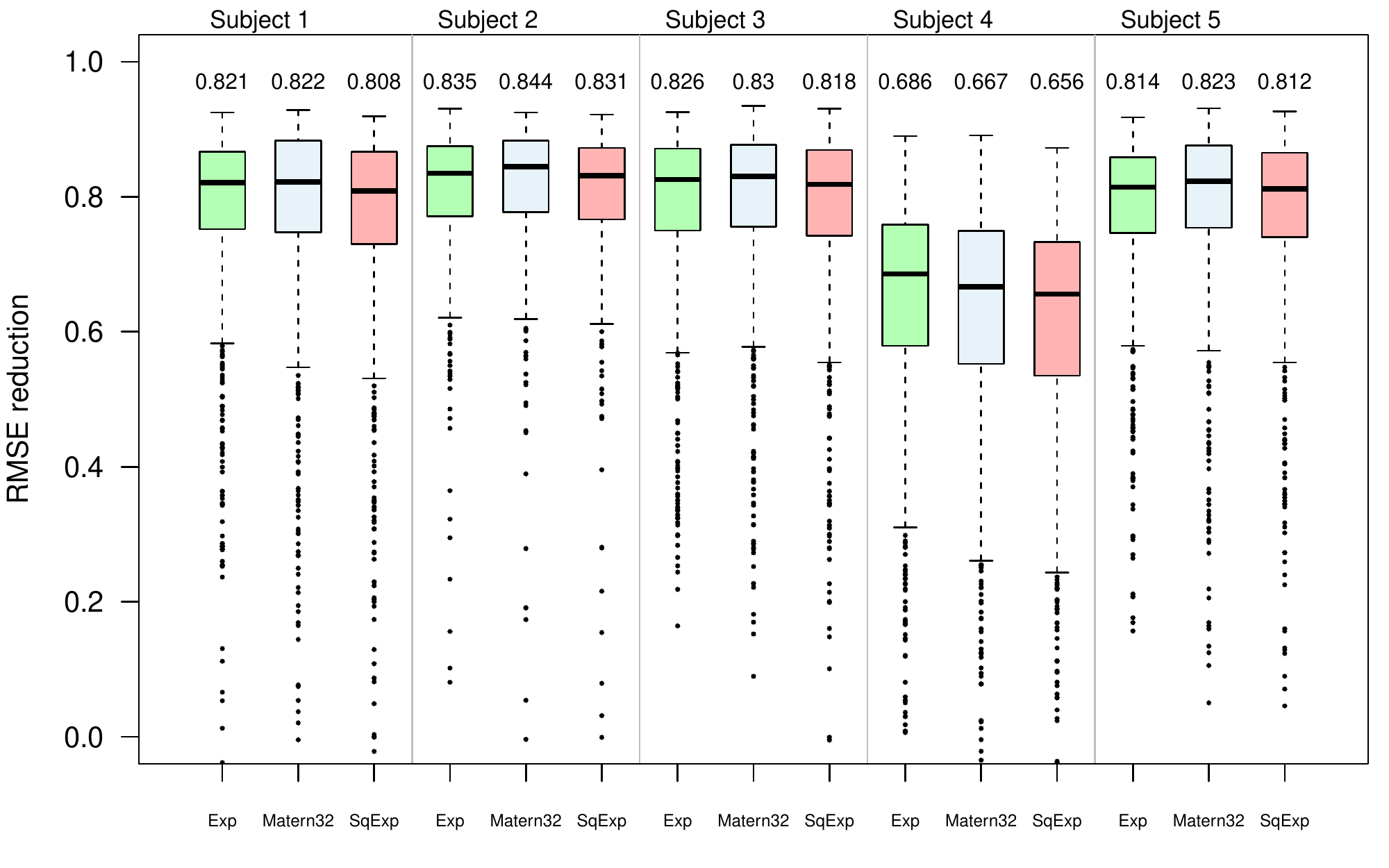}
    \caption{Boxplots of RMSE reduction when using exponential (Exp) covariance (green) and Mat\'ern $\nu = 1.5$ (Mat32) covariance (lightblue), and squared exponential (sqExp) covariance (pink) with standardized harmonic representation features and lag map.}
    \label{fig:figS5}
\end{figure}

\section{Number of nearest-neighbors}

\begin{figure}[H]
    \centering
    \includegraphics[width=4in]{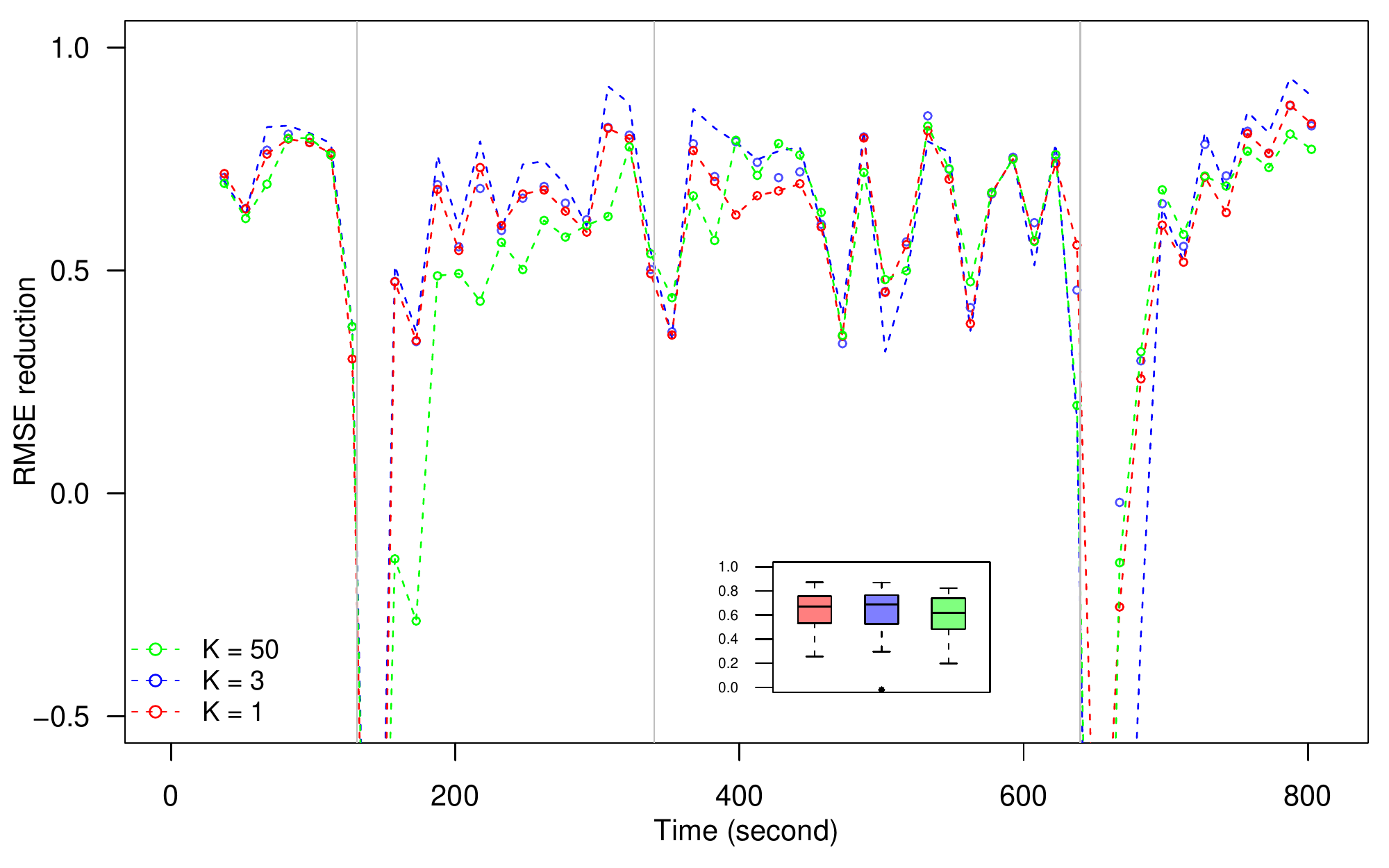}
    \caption{RMSE reduction across all 15-second windows with different $K$ when fitting \texttt{LocGP}. The number of NNs considered are $1$ (red), $3$ (blue), and $50$ (green).}
    \label{fig:figS6}
\end{figure}

When fitting both \texttt{LocGP} and \texttt{LocDBGP}, one of the tuning parameters is $K$, the number of NNs used to form the training set. A sensitivity analysis by letting $K= 1$, $3$, and $50$ is carried out on the UPenn dataset. Fig.~\ref{fig:figS6} indicates that the prediction performance is relatively insensitivity with respect to $K$, except for the first half part of segment 2, where the prediction using $K=50$ is worse than that of the predictions with $K=1$ and $3$. This result is not surprising. Since there are few epochs during the transition period, it is better to consider a small $K$. The results here also suggest that $K=3$ is a reasonable choice.     

\begin{figure}[H]
    \centering
    \includegraphics[width=4in]{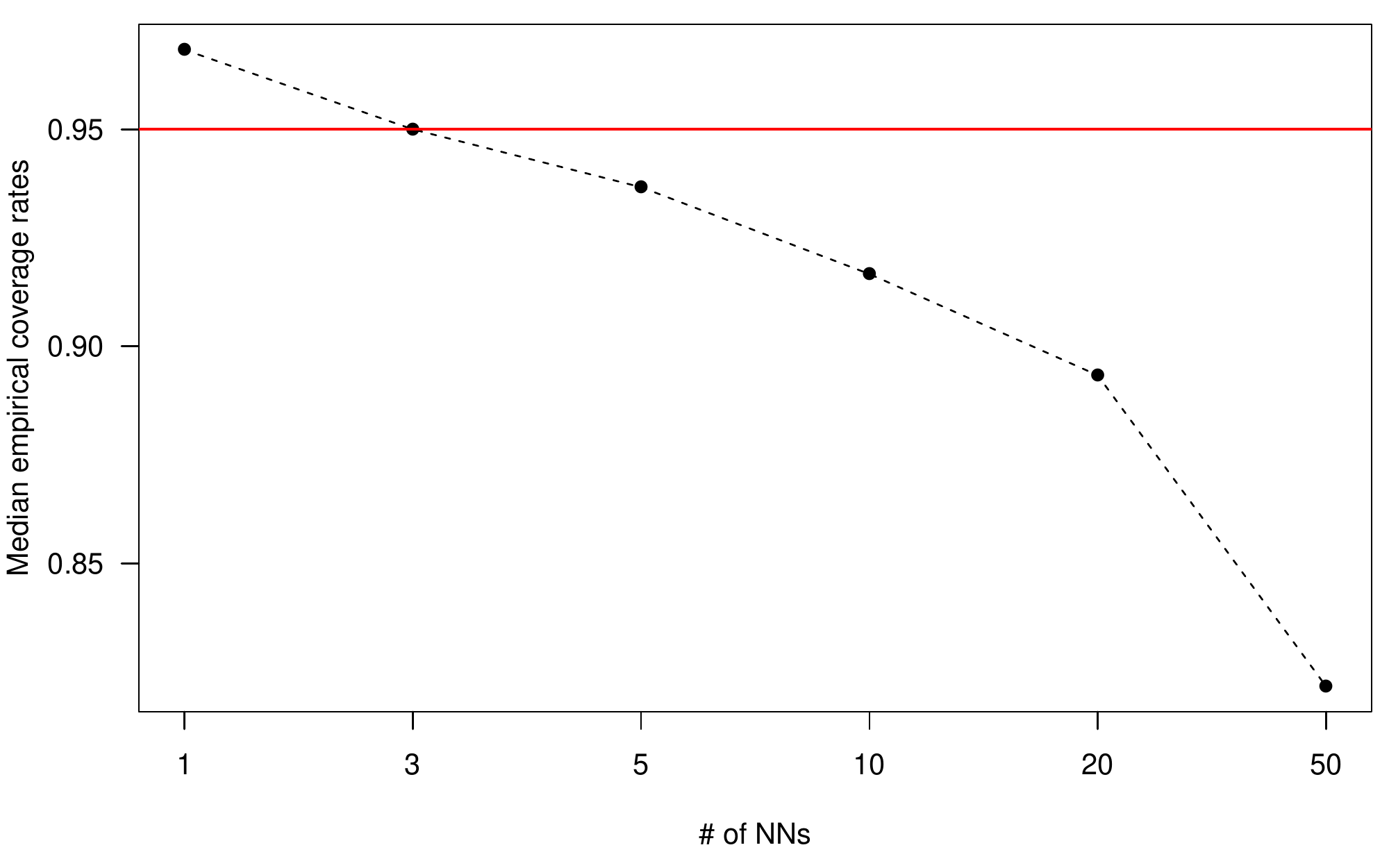}
    \caption{The medians of empirical coverage rates of the estimated point-wise 95\% confidence interval for all the 30-second time windows with a range of $K$.}
    \label{fig:figS7}
\end{figure}

We also examine how the empirical coverage rate (ECP) depends on the number of NNs used. Fig.~\ref{fig:figS7} suggests that the ECP might decrease with $K$, where the higher than the nominal rate (i.e., 0.95) with $K=1$ may suggest the interval is too wide and therefore the inference is not ``sharp''. 

\section{TIDIS sleep stage information}

Table. 1 provides the relative frequency of each sleep stage for each subject included in the TIDIS dataset. It is worth pointing out that Subject 4 has a relatively high proportion of awake stage.  

\begin{table}[H]
\begin{center}
\begin{tabular}{rrrrrr}
  \hline
 & Subject 1 &Subject 2 &Subject 3 &Subject 4 &Subject 5 \\ 
  \hline
N3 & 0.23 & 0.39 & 0.18 & 0.02 & 0.17 \\ 
  N2 & 0.46 & 0.34 & 0.45 & 0.42 & 0.67 \\ 
  N1 & 0.06 & 0.06 & 0.07 & 0.15 & 0.02 \\ 
  REM & 0.19 & 0.17 & 0.27 & 0.07 & 0.10 \\ 
  W & 0.06 & 0.04 & 0.02 & 0.35 & 0.04 \\ 
   \hline
\end{tabular}
\end{center}
\caption{The relative frequency for each sleep stage for each TIDIS subject.}
\end{table}

Table. 2 shows the transition proportion for each subject. Subject 4 has the highest transition proportion.

\begin{table}[H]
\begin{center}
    \begin{tabular}{lrrrrr}
  \hline
 &Subject 1 &Subject 2 &Subject 3 &Subject 4 &Subject 5 \\ 
  \hline
Transition proportion  & 0.11 & 0.08 & 0.11 & 0.24 & 0.06 \\ 
\# of time window & 835 &732& 878& 901 &761\\ 
   \hline
\end{tabular}
\end{center}
\caption{The transition proportion and the number of 30-second time windows that sleep stage were recorded for each subject.}
\end{table}

Table. 3 gives the relative frequency of each sleep stage during these transition time windows. It is worth pointing out that the sleep stage N1 has much higher tendency, compared with all other sleep stages, being a transition stage.

\begin{table}[H]
\begin{center}
    \begin{tabular}{rrrrrr} 
  \hline
 &Subject 1 &Subject 2 &Subject 3 &Subject 4 &Subject 5 \\ 
  \hline
N3 & 0.10 & 0.14 & 0.06 & 0.01 & 0.13 \\ 
  N2 & 0.35 & 0.39 & 0.45 & 0.42 & 0.40 \\ 
  N1 & 0.35 & 0.25 & 0.41 & 0.38 & 0.31 \\ 
  REM & 0.08 & 0.12 & 0.07 & 0.01 & 0.09 \\ 
  W & 0.10 & 0.09 & 0.01 & 0.17 & 0.07 \\ 
   \hline
\end{tabular}
\end{center}
\caption{The relative frequency for each sleep stage conditioning during transition periods for each TIDIS subject.}
\end{table}

\section{TIDIS within-subject RMSE}
Due to the space limitation in the main article, we only provide the RMSE reduction time series for all the consecutive 30-second windows for Subject 2. Here we plot the RMSE reduction time series for each subject.

\begin{figure}[bhp!]
    \centering
    \includegraphics[page =1,width=4.5in]{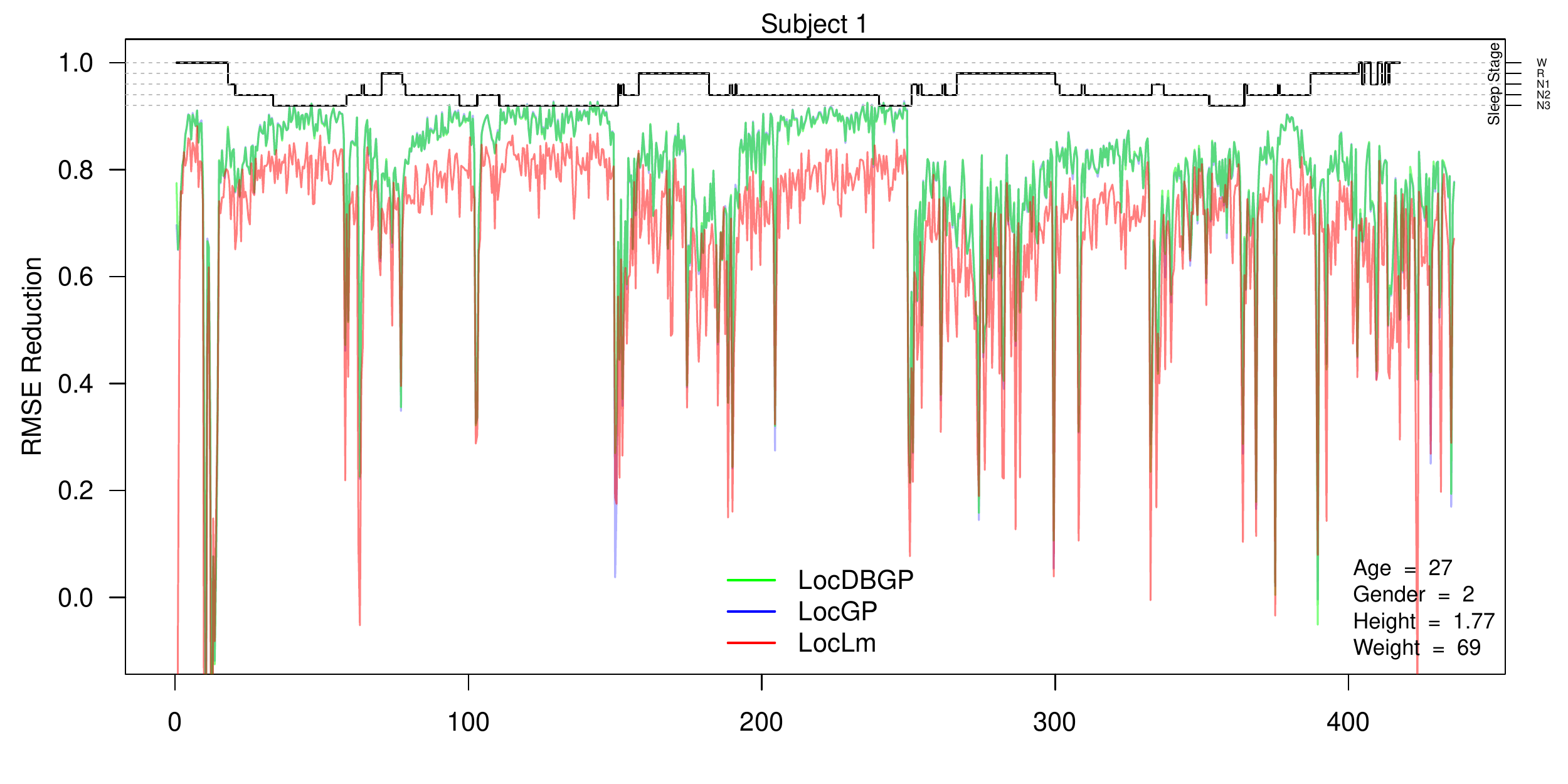}\\
      \includegraphics[page =2,width=4.5in]{Fig/NNGPInRMSE_suppV2.pdf}\\
      \includegraphics[page =3,width=4.5in]{Fig/NNGPInRMSE_suppV2.pdf}
    \caption{The time series of the RMSE reduction for all the consecutive 30-second windows for Subjects 1, 2, and 3 with sleep stage information included on the top. Note that the RMSE between \texttt{LocDBGP} and \texttt{LocGP} are almost indistinguishable and therefore only the RMSEs for \texttt{LocDBGP} shown here.}
    \label{fig:figS8}
\end{figure}

\begin{figure}[bhp!]
    \centering
      \includegraphics[page =4,width=4.5in]{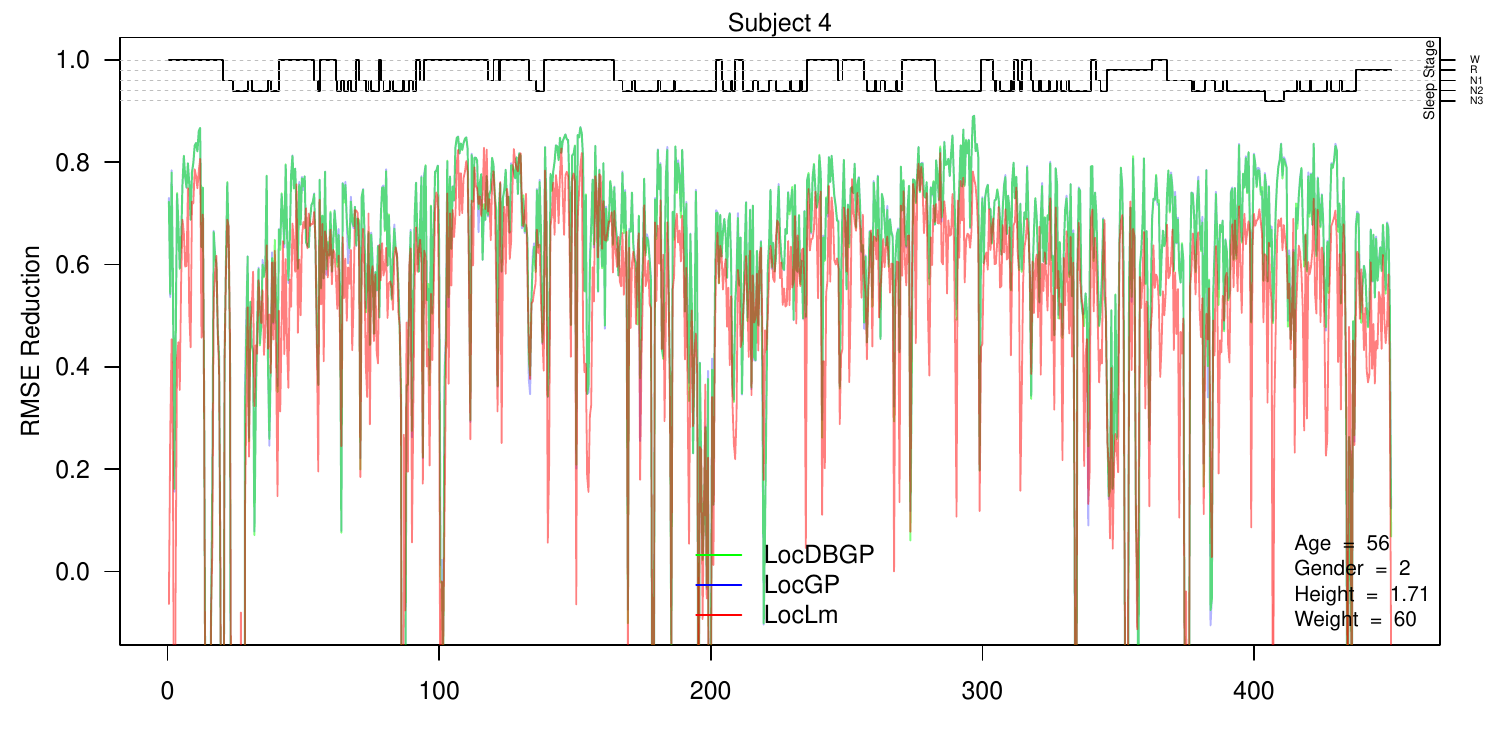}\\
      \includegraphics[page =5,width=4.5in]{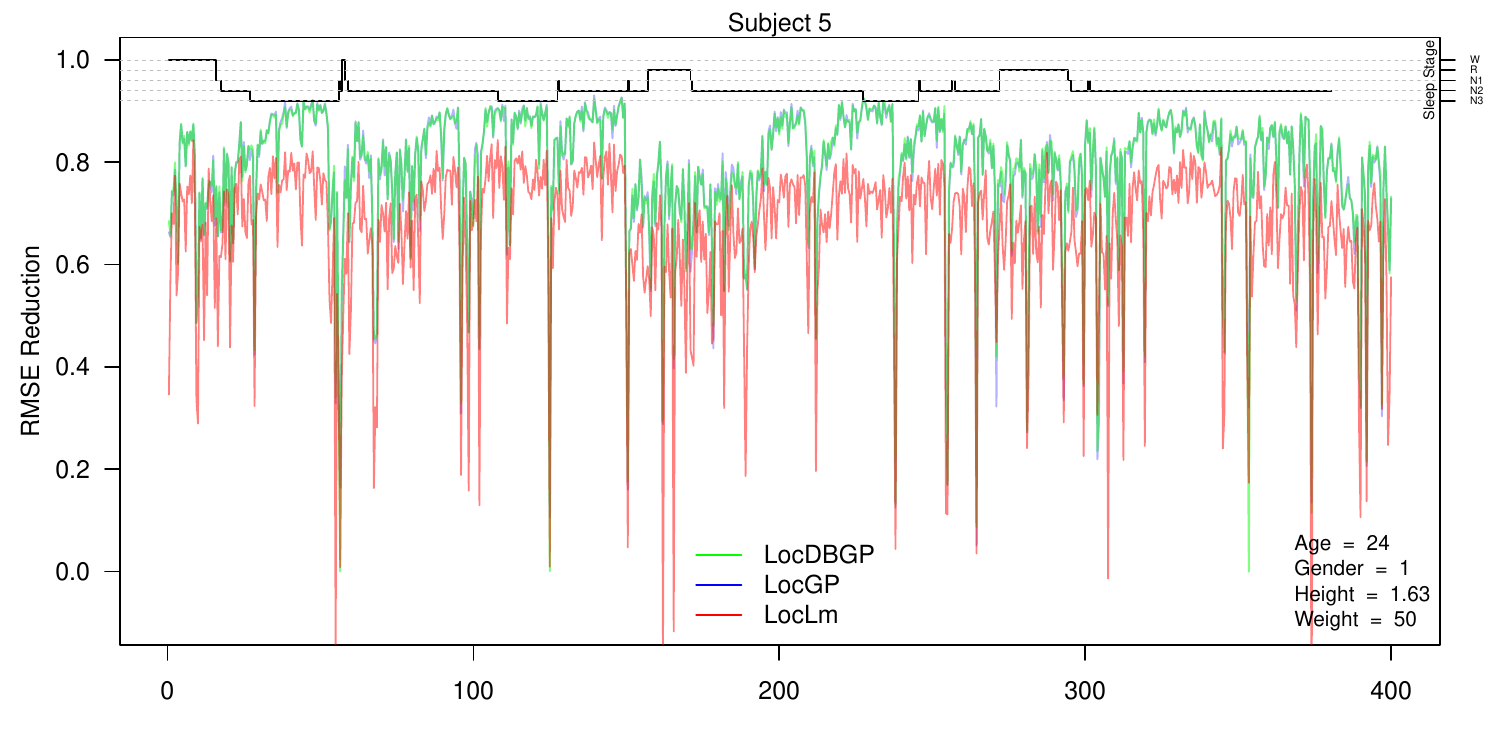}
    \caption{As in Fig.~\ref{fig:figS6} but for subjects 4, and 5.}
    \label{fig:figS9}
\end{figure}

\section{TIDIS inter-subject RMSE}

Due to the space limitation in the main article, we only provide the RMSE reduction time series for all the consecutive 30-second windows for Subject 3. Here we plot the RMSE reduction time series for Subject 4, 5.

\begin{figure}[bhp!]
    \centering
      \includegraphics[page =2,width=5in]{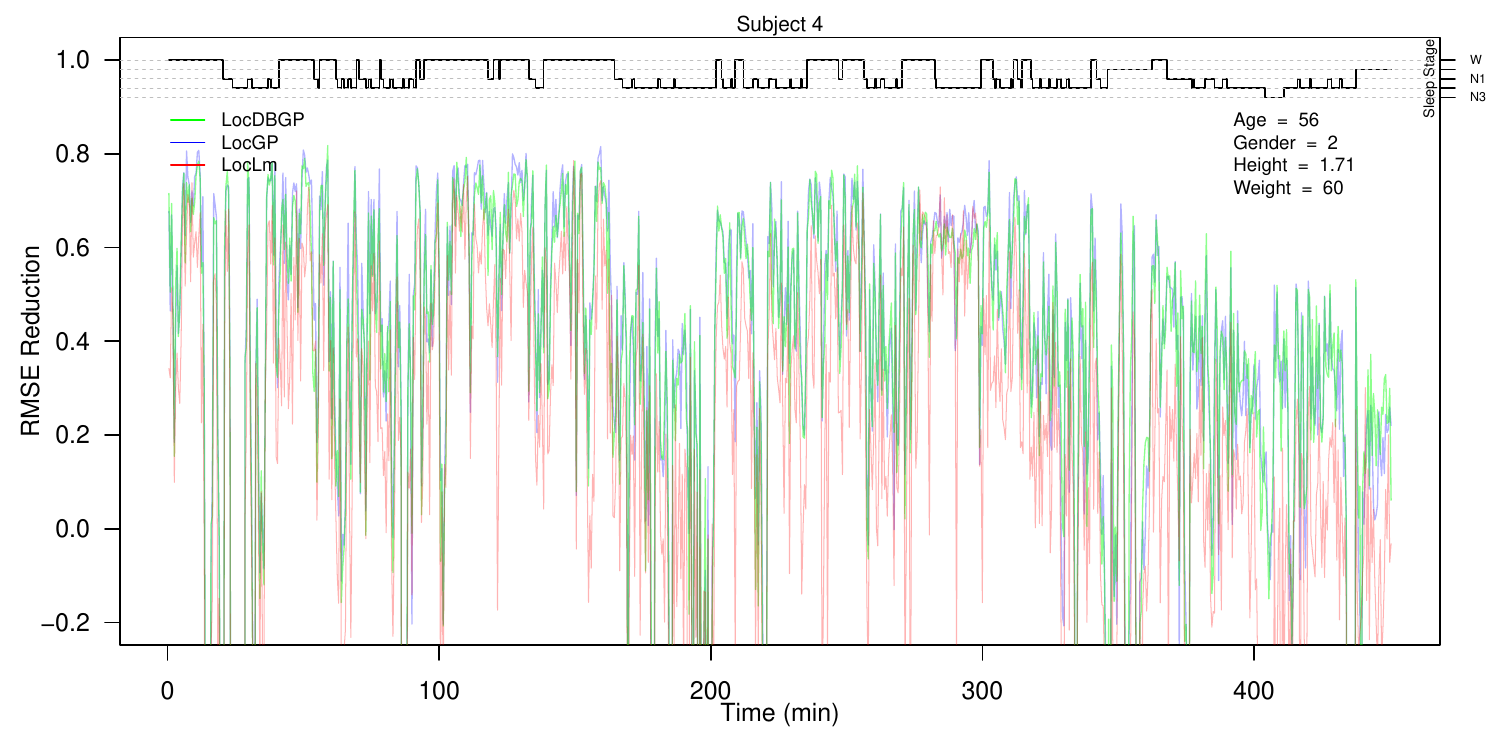}\\
      \includegraphics[page =3,width=5in]{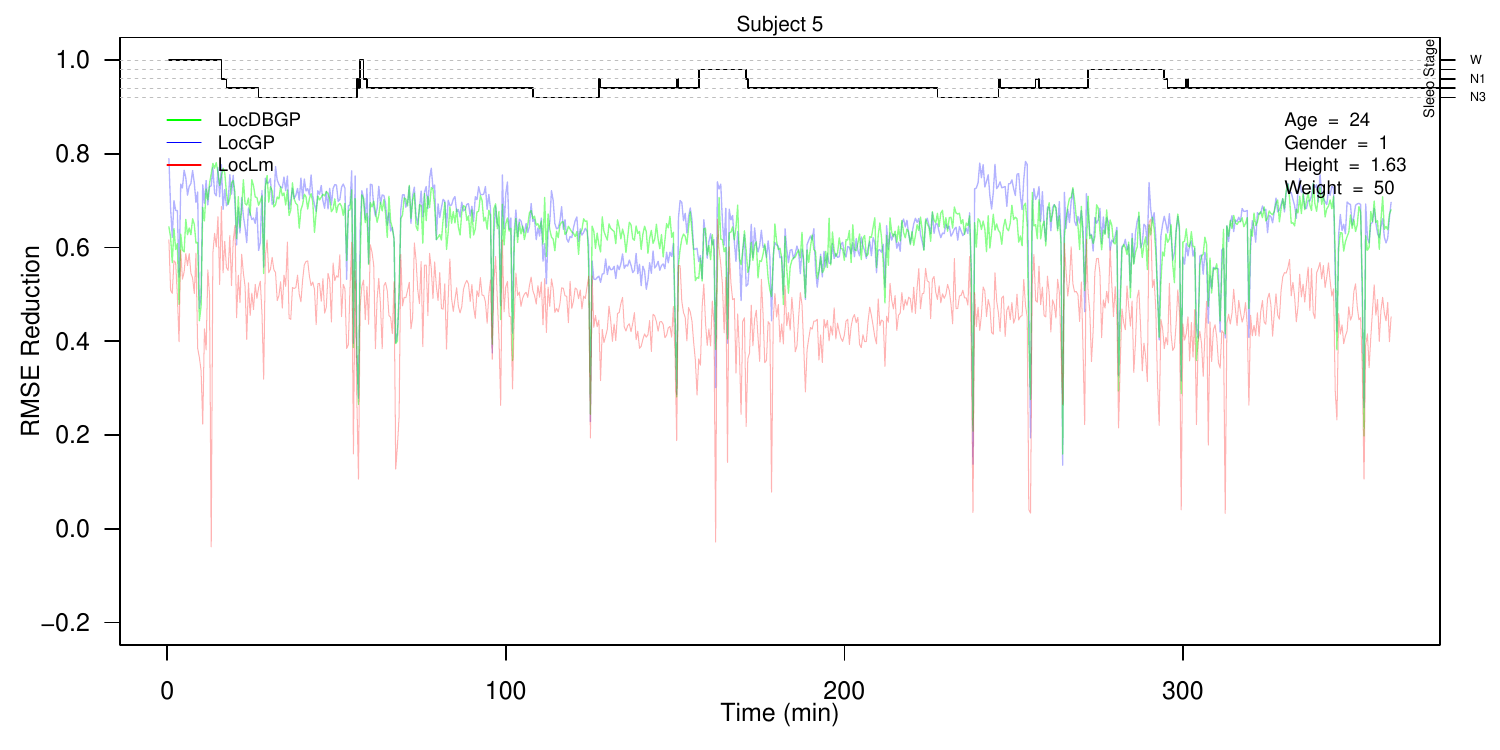}
    \caption{The time series of the RMSE reduction (plotted every 30 seconds) for Subjects 4, and 5 with sleep stage information included on the top.}
    \label{fig:figS10}
\end{figure}

\section{TIDIS inter-subject prediction RMSE with all the combinations of two training subjects}

Here we present the RMSE reduction for the inter-subject prediction of each subject using any two other subjects as the training data so there $\binom{4}{2} \times 5 =6 \times 5 = 30$ different combinations. 
As explained in the main text, here we apply global alignment to account for possible phase shifts. For most cases these shifts are relatively small (with 0.2 sec) except for all the cases for Subject 5 where the phase shifts are either 0.9 or 1.0 second backward and for Subject 2 using Subject 4 and 5 as the training set where the phase is 0.9 second forward. The number of NNs used here is 1 rather than 3.        

\begin{figure}[bhp!]
    \centering
    \includegraphics[page = 1,width=3.8in]{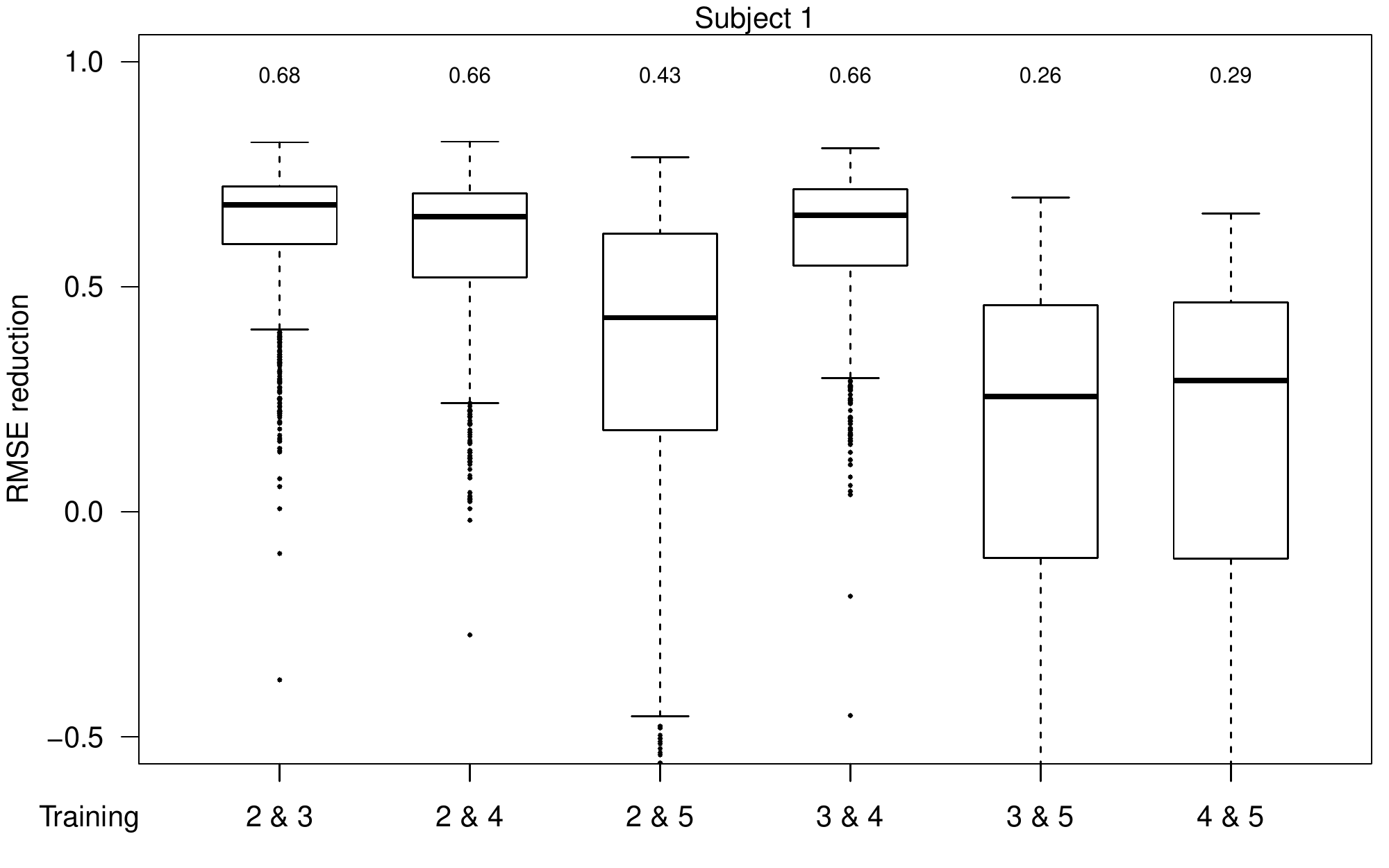}\\
      \includegraphics[page =2,width=3.8in]{Fig/OutRMSEAll.pdf}\\
      \includegraphics[page =3,width=3.8in]{Fig/OutRMSEAll.pdf}
    \caption{Boxplots of RMSE reduction for Subjects 1, 2, and 3 with different training set (e.g., 2 \& 3 means the predictions are obtained using Subjects 2 and 3 as the training data).}
    \label{fig:figS11}
\end{figure}

\begin{figure}[bhp!]
    \centering
    \includegraphics[page = 4,width=4in]{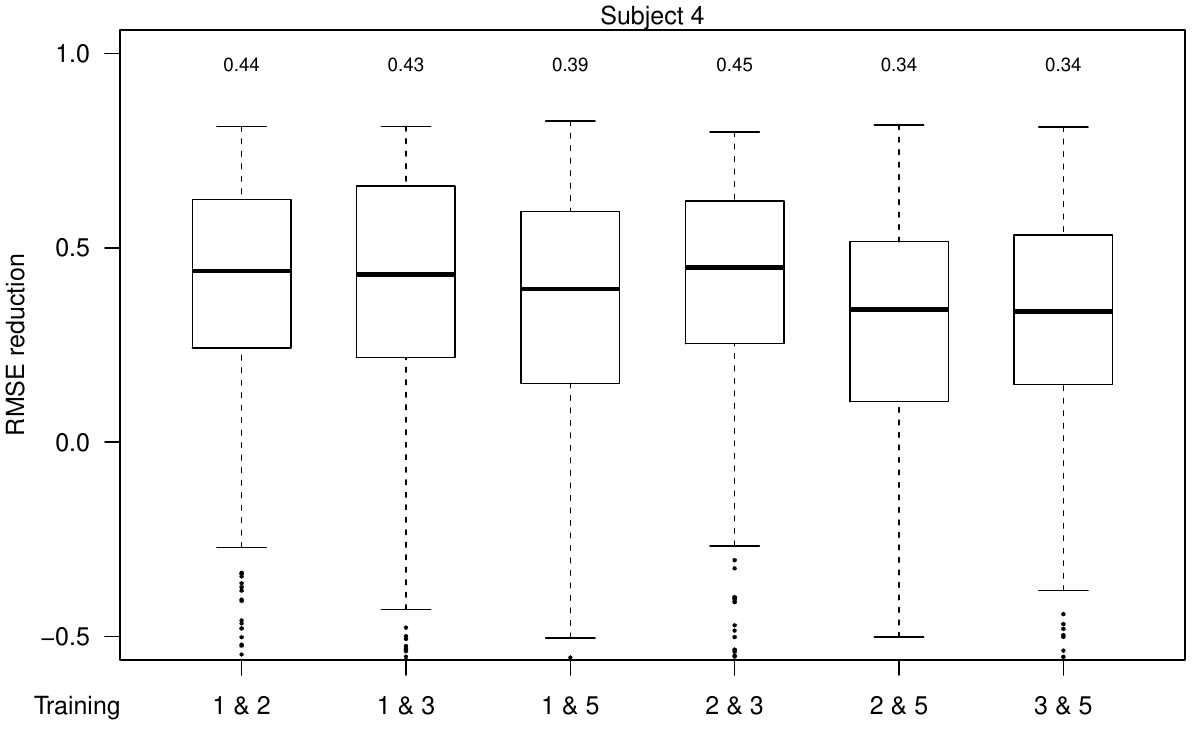}\\
      \includegraphics[page =5,width=4in]{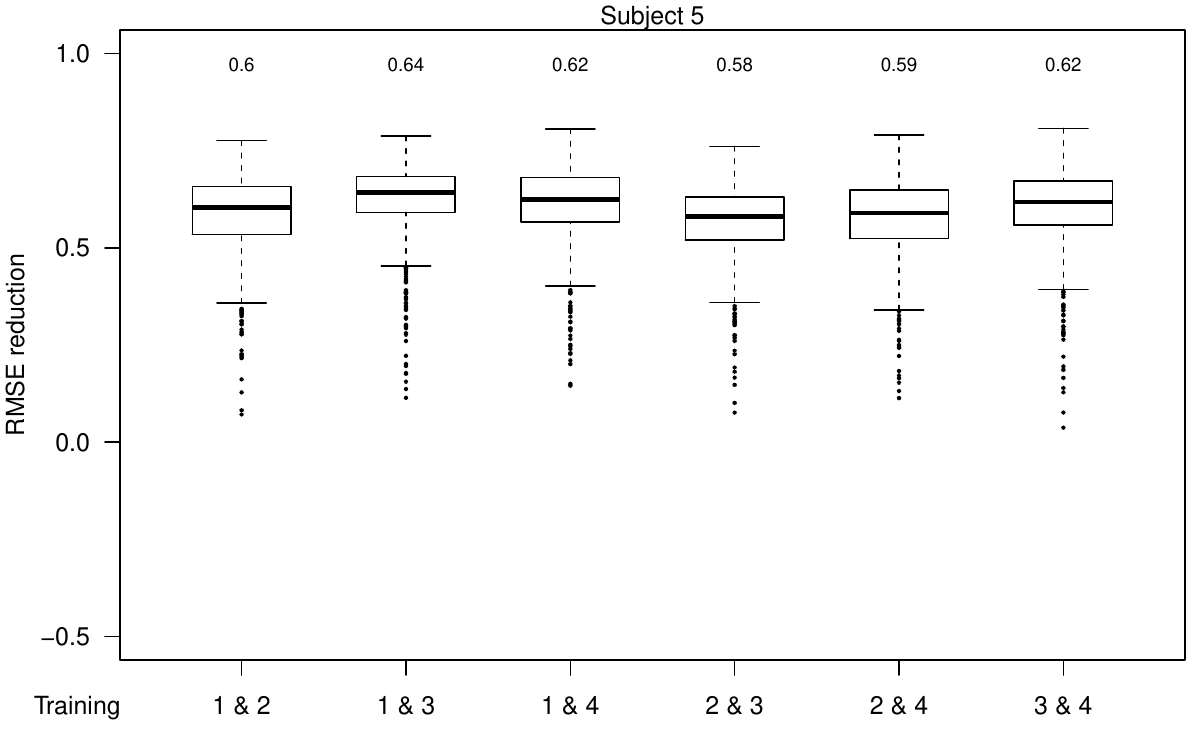}
    \caption{As in Fig.~\ref{fig:figS11} but for Subjects 4 and 5.}
    \label{fig:figS12}
\end{figure}

\end{document}